\documentclass[letterpaper, 10 pt, conference]{ieeeconf}  % Comment this line out if you need a4paper

\IEEEoverridecommandlockouts                              % This command is only needed if 
\usepackage{graphicx} % for pdf, bitmapped graphics files
\usepackage{amsmath} % assumes amsmath package installed
\usepackage{amssymb}  % assumes amsmath package installed
\usepackage{mathtools}
\usepackage[noadjust]{cite}
\usepackage{booktabs}
\usepackage{hyperref}
\usepackage{cleveref}
\usepackage{siunitx}
\usepackage{tikz}
\usetikzlibrary{patterns, shapes.geometric, arrows.meta, fit, 3d}
\usepackage{pgfplots}
\usepackage{textcomp}
\usepackage{color,soul}

\makeatletter
\let\MYcaption\@makecaption
\makeatother

\usepackage[font=footnotesize]{subcaption}

\makeatletter
\let\@makecaption\MYcaption
\makeatother

\crefname{figure}{fig.}{Fig.}
\renewcommand{\cref}{\Cref}

\makeatletter
\tikzoption{canvas is xy plane at z}[]{%
	\def\tikz@plane@origin{\pgfpointxyz{0}{0}{#1}}%
	\def\tikz@plane@x{\pgfpointxyz{1}{0}{#1}}%
	\def\tikz@plane@y{\pgfpointxyz{0}{1}{#1}}%
	\tikz@canvas@is@plane
}
\makeatother

\tikzset{xyp/.style={canvas is xy plane at z=#1}}
\tikzset{xzp/.style={canvas is xz plane at y=#1}}
\tikzset{yzp/.style={canvas is yz plane at x=#1}}

\newcommand{\mynorm}[1]{\left\lVert#1\right\rVert}

\newcommand\centerofmass{%
	\tikz[radius=0.4em] {%
		\fill (0,0) -- ++(0.4em,0) arc [start angle=0,end angle=90] -- ++(0,-0.8em) arc [start angle=270, end angle=180];%
		\draw (0,0) circle;%
	}%
}

\pgfdeclarepatternformonly{south east lines}{\pgfqpoint{-0pt}{-0pt}}{\pgfqpoint{6pt}{6pt}}{\pgfqpoint{6pt}{6pt}}{
	\pgfsetlinewidth{0.4pt}
	\pgfpathmoveto{\pgfqpoint{0pt}{3pt}}
	\pgfpathlineto{\pgfqpoint{3pt}{0pt}}
	\pgfpathmoveto{\pgfqpoint{2.5pt}{-2.5pt}}
	\pgfpathlineto{\pgfqpoint{-2.5pt}{2.5pt}}
	\pgfpathmoveto{\pgfqpoint{5.5pt}{0.5pt}}
	\pgfpathlineto{\pgfqpoint{0.5pt}{5.5pt}}
	\pgfusepath{stroke}}

\pgfdeclarepatternformonly{south west lines}{\pgfqpoint{-0pt}{-0pt}}{\pgfqpoint{15pt}{15pt}}{\pgfqpoint{15pt}{15pt}}{
	\pgfsetlinewidth{0.4pt}
	\pgfpathmoveto{\pgfqpoint{0pt}{0pt}}
	\pgfpathlineto{\pgfqpoint{15pt}{15pt}}
	\pgfpathmoveto{\pgfqpoint{14.8pt}{-.2pt}}
	\pgfpathlineto{\pgfqpoint{15.2pt}{.2pt}}
	\pgfpathmoveto{\pgfqpoint{-.2pt}{14.8pt}}
	\pgfpathlineto{\pgfqpoint{.2pt}{15.2pt}}
	\pgfusepath{stroke}}

\title{\LARGE \bf
Zero-shot sim-to-real transfer of tactile control policies \\for aggressive swing-up manipulation %(or maneuvers?)
}

\author{Thomas Bi, Carmelo Sferrazza and Raffaello D'Andrea%\\[1em]% <-this % stops a space
\thanks{The authors are members of the Institute for Dynamic Systems and Control, ETH Zurich, Switzerland. Email correspondence to Thomas Bi
        {\tt\small bit@ethz.ch}}%
}

\renewcommand{\hl}{}

\begin{document}

\maketitle
\thispagestyle{empty}
\pagestyle{empty}

%%%%%%%%%%%%%%%%%%%%%%%%%%%%%%%%%%%%%%%%%%%%%%%%%%%%%%%%%%%%%%%%%%%%%%%%%%%%%%%%
\begin{abstract}
	
%Tactile sensors provide robots with a sense of touch akin to the one provided by human fingertips. 
This paper aims to show that robots equipped with a vision-based tactile sensor can perform dynamic manipulation tasks without prior knowledge of all the physical attributes of the objects to be manipulated. For this purpose, a robotic system is presented that is able to swing up poles of different masses, radii and lengths, to an angle of 180\textdegree, while relying solely on the feedback provided by the tactile sensor. This is achieved by developing a novel simulator that accurately models the interaction of a pole with the soft sensor. A feedback policy that is conditioned on a sensory observation history, and which has no prior knowledge of the physical features of the pole, is then learned in the aforementioned simulation. 
When evaluated on the physical system, the policy is able to swing up a wide range of poles that differ significantly in their physical attributes without further adaptation. To the authors' knowledge, this is the first work where a feedback policy from high-dimensional tactile observations is used to control the swing-up manipulation of poles in closed-loop.

\end{abstract}

%%%%%%%%%%%%%%%%%%%%%%%%%%%%%%%%%%%%%%%%%%%%%%%%%%%%%%%%%%%%%%%%%%%%%%%%%%%%%%%%

\section{INTRODUCTION}
\label{sec:intro}

%The sense of touch provides humans with the dexterity needed to handle objects of various shapes and allows them to expertly perform intricate tasks. For example, we can spin pens around our fingers, rotate a spinning basketball on our index finger, or swing up a pen in our hands. Often, we possess the ability to accomplish such tasks without relying on our sense of vision, using the feedback provided by the tactile receptors in our skin instead. Moreover, a change in the properties of the object to be manipulated, typically does not alter our success. A change in weight of the pen, for instance, does not deter us from swinging it up to the desired angle.

\begin{figure}%[!h]
	\centering
	\begin{tikzpicture}
		\node at (0, -2.3)
		{\includegraphics[width=0.83\columnwidth]{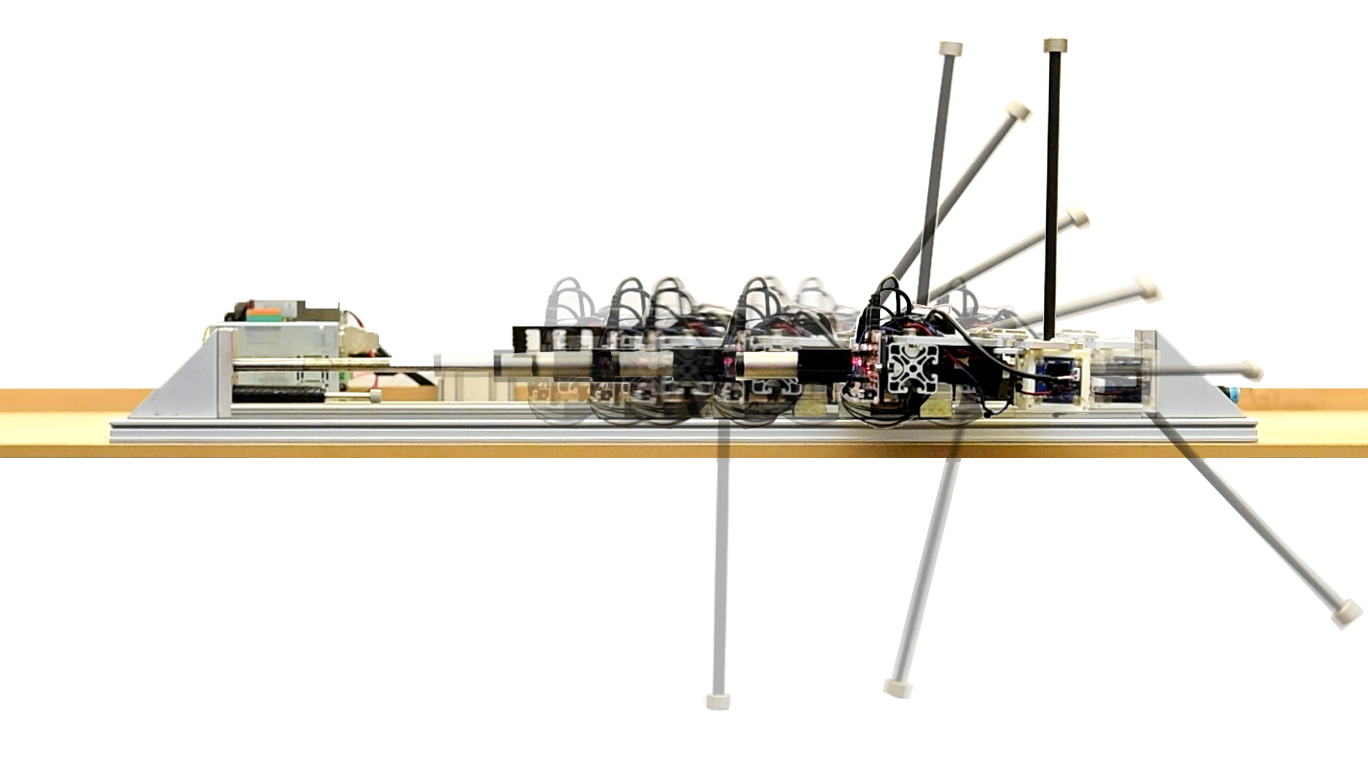}};
		
		\node at (0, 1.5)
		{\includegraphics[width=0.83\columnwidth]{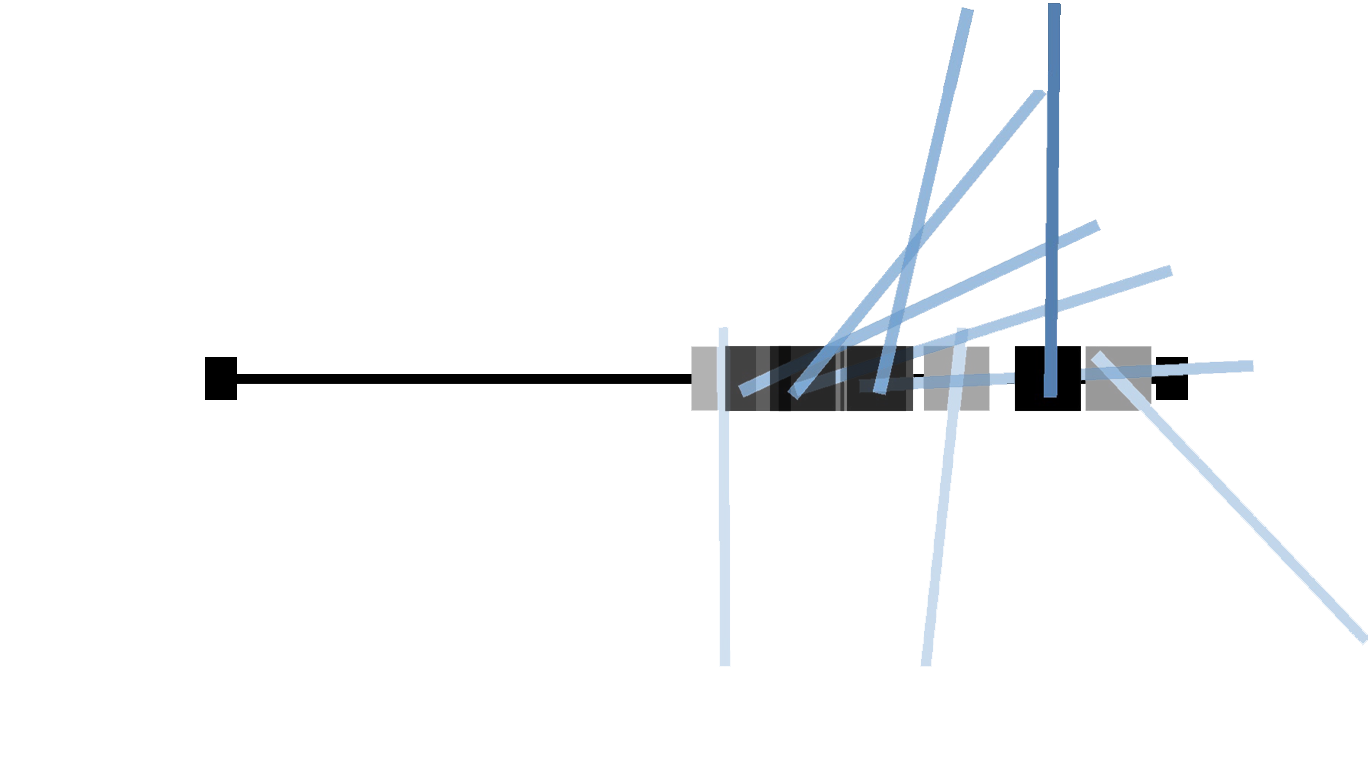}};

		\node at (0, -0.3) {\small (a) Simulation};
		\node at (0, -4.3) {\small (b) Reality};
		
	\end{tikzpicture}
	
	\caption{In this work, tactile control policies for the swing-up manipulation of poles are learned in a simulation of the tactile sensor that runs faster than real-time. When transferred to the real-world system, the policy achieves the desired behaviour without further adaptation, and swings up poles to an upright position without prior knowledge of the physical attributes of the pole.}
	\vspace{-0.5cm}
	\label{fig:intro}
\end{figure}

Tactile sensors aim to provide robots with a sense of touch %a similar sense of touch to robots, as % ... deformable nature of soft sensor makes modeling and control hard..,
that captures information from their environment through physical contact. In this paper, the vision-based tactile sensor presented in \cite{sferrazza2019design} is deployed in order to demonstrate that it can provide robots with a dexterity akin to that of humans in dynamic manipulation tasks. %, thereby showcasing that the sense of touch plays a vital part in solving unstructured and complex dexterous manipulations tasks. Following ...capabilities ...
For this purpose, a robotic system that performs swing-up maneuvers for different poles is presented (see \cref{fig:intro}). The robotic system consists of a parallel gripper, mounted to a linear motor, with two tactile sensors acting as fingers. 
Thereby, three key capabilities enabled by the artificial sense of touch provided by the tactile sensor are demonstrated: (i) The system is able to adapt its motion and successfully swings up poles that differ in their physical attributes (e.g. mass, length, and radius) without prior knowledge of these  attributes. (ii) The system does not rely on external visual sensing; instead, the pose and attributes of the pole in contact are implicitly inferred from the tactile observations alone. (iii) The tactile observations can be processed in real-time and act as feedback for closed-loop control at $\SI{60}{\hertz}$. As a result, highly dynamic swing-up manipulations are achieved without the need for a previous in-hand exploration of the pole.

The three components that enable such adaptive dynamic swing-up manipulation are presented here. First, the high-dimensional force distribution acting on the sensor surface is directly inferred from the sensor camera images using an efficient convolutional network, which is trained on purely simulated contact interactions of the sensor with different poles. Second, a novel simulator is developed that accurately models the behaviour of the soft sensor surface when interacting with a rigid cylindrical pole. This simulation is based on combining the finite element method with state-of-the-art semi-implicit time-stepping schemes for contact resolution and runs at $\SI{360}{\hertz}$ on a single core of an Intel Core i7-7700k processor. Third, a framework for learning adaptive feedback policies conditioned on a history of sensory observations is proposed. The training of the policy is achieved entirely in simulation using deep reinforcement learning. Thereby, various strategies that facilitate the sim-to-real transfer of policies learned in simulation are employed, namely \emph{dynamics randomization} \cite{peng2018sim}, and \emph{privileged learning} \cite{chen2020learning, lee2020learning}.% {\color{red} more detailed description of what exactly we do?}% and the various strategies that have been presented for the sim-to-real tra%two steps. First, given the promising capabilities of deep reinforcement learning to learn desired control policies in simulation... Moreover, dynamics randomization.. 

%Since the parameters and pose of the pole are privileged information which are not directly observable on the real system, 

%Two steps..%a framework that allows for a sim-to-real transfer of control polcies learned in simulation is proposed.
%Third, we propose

%Our main contribution can be divided into three parts.
 %Second, we present a fast simulator that accurately models the behaviour of the soft sensor surface when interacting with a pole. This simulation can be run at $\SI{360}{\hertz}$.  One such policy is trained to swing up poles to an angle of $\SI{180}{\degree}$. Experiments on the real system show that the policy successfully swings up poles of various different physical properties, e.g. different radii, masses and lengths. Moreover, the dynamics of the simulator are shown to closely match the behaviour of the real system. Additionally, as proof of concept of the generalizability of our approach, we show that policies for other tasks, such as a throw-and-catch maneuver of the pole, can be learned in simulation.

\subsection{Related Work}

As reviewed in \cite{yousef2011tactile} and \cite{kappassov2015tactile},
\hl{many works have demonstrated how robots can leverage the sense of touch in dexterous manipulation tasks in closed-loop. For example, in {\cite{romano2011human}} tactile data is used to control both the grasping force and slippage of a tactile gripper. Other examples include {\cite{she2019cable}}, where a pair of grippers are used to pick one end of a cable and follow it to the other end. Each gripper contains tactile sensors from which the current pose and friction forces acting on the cable can be estimated in real-time,} %Using a learned dynamics model from data collected on the physical system, the cable following is then performed in closed-loop using the estimated pose and friction forces as feedback.
enabling the approach to generalize to cables of different thicknesses and materials, based on a model learned from real data. A similar approach was proposed in \cite{hogan2020tactile}; a dual palm robotic system estimates the pose and the stick/slip behaviour of an object solely from tactile feedback, in order to manipulate an object on a planar surface to a desired position. In \cite{tian2019manipulation}, a deep dynamics model is learned that can predict future tactile observations based on the previous observations and actions taken. Data for the training of the dynamics model is autonomously collected on the physical system. The learned model is then used in an MPC-framework to manipulate a ball, analog stick, and 20-sided die to a desired configuration. Other learning-based approaches rely on deep reinforcement learning to %explore and 
find optimal control policies directly on the physical hardware. Examples include a 5-DoF arm that learns to reorient objects using a latent representation of the tactile data \cite{van2016stable}, and a robotic system that learns to type on a Braille keyboard \cite{church2020deep}.

While these approaches demonstrate robustness against external disturbances and changes in object properties, the manipulation tasks they solve generally do not require a high degree of dynamicism. For more aggressive manipulation tasks such as the swing-up manipulation of poles, feedback control based on tactile data has proven to be challenging and thus differing methods have been proposed. %For example,  
In \cite{wangswingbot}, the sensing and manipulation are separated into two steps. First, the physical features of different poles are learned by shaking and tilting the pole in-hand and observing the tactile feedback.
The learned features are then used to optimize an open-loop trajectory of a robotic arm that dynamically swings the pole up to a desired angle. The learning of the physical features, as well as the trajectory optimization, are performed end-to-end using models trained on a physically collected dataset.
In \cite{adaptivepivoting}, tactile sensing and visual tracking are combined to pivot an object to a desired angle by adjusting the gripping force exerted by a two-finger gripper. This fusion of visual and tactile information was also employed on a robotic hand in \cite{senoo2009skillful} to perform highly dynamic tasks such as pen spinning, ball dribbling, and ball throwing.
%The ..Skillful manipulation based on high-speed sensory-motor fusion \cite{senoo2009skillful}
%Other methods have been proposed for the swing-up of objects grasped by a gripper.
%Hou et al. \cite{robustpivoting} proposed that planar in-hand pivoting of an object can be achieved simply by controlling the gripping force acting on the object. Similarly, in \cite{roboticswingup} the normal force on a grasped object and the motion of a robotic arm are controlled in a way to swing up the object to a desired angle.

%Often, ..high-dimensional observations are condensed to user-designed, intermittent features, such as object pose and slippage. In contrast, 

%Tactile-Based Insertion for Dense Box-Packing

%Dynamic pen spinning using a high-speed multifingered hand with high-speed tactile sensor

%Event-Driven Visual-Tactile Sensing and Learning for Robots \cite{taunyazov2020event}

%Exploratory Tactile Servoing With Active Touch

%The mentioned learning 
%Simulation here??

In this work, a unified approach is presented where aggressive swing-up maneuvers can be achieved in closed-loop from high-dimensional tactile feedback, without relying on a visual tracking system or prior in-hand exploration of the pole. Moreover, instead of relying on data collection on the physical system, as is done in the learning-based methods mentioned above, the feedback control policy is learned entirely in simulation. This removes the cost of collecting data on the physical system, which can be highly time-consuming. Additionally, challenging motions that may lead to unsafe behaviours by the physical hardware can be first explored without repercussions. This data can then be utilized to 
train the policy to satisfy the safety constraints that are present on the physical system. While simulators for the behaviour of tactile sensors have been developed (see e.g. \cite{ding2020sim, wang2020elastic, kappassov2020simulation, joergensen2010robworksim, moisio2013model, bauza2020tactile, softrob, sferrazza2020learning}), the authors are not aware of any work where a simulation from first principles is utilized to learn tactile feedback control policies. Rather, the mentioned works focus on gathering supervised datasets of tactile images in simulation to train deep neural networks that can predict object position and rotation (\cite{ding2020sim, bauza2020tactile}), the force distribution acting on the sensor surface (\cite{softrob, sferrazza2020learning}), or the three-dimensional mesh of the object in contact (\cite{wang2020elastic}).

\subsection{Outline}

The hardware employed for the experiments is presented in \cref{sec:hardware}. In \cref{sec:method}, the proposed methods are described. This includes the sensing approach of the tactile sensor in \cref{sec:sens}, the design of the tactile simulator in \cref{sec:sim}, and the synthesis of the swing-up control policy in \cref{sec:control}. Results from employing the learned policy on the real-world system are presented in \cref{sec:results}. Finally, \cref{sec:conc} draws conclusions and gives an outlook on future work. In the remainder of this paper, vectors are expressed as tuples for ease of notation, with dimension and stacking clear from the context.

\section{HARDWARE}
\label{sec:hardware}

\begin{figure}
	\centering
	\begin{tikzpicture}
		\node[inner sep=0pt] at (0, 0) {\includegraphics[width=\columnwidth]{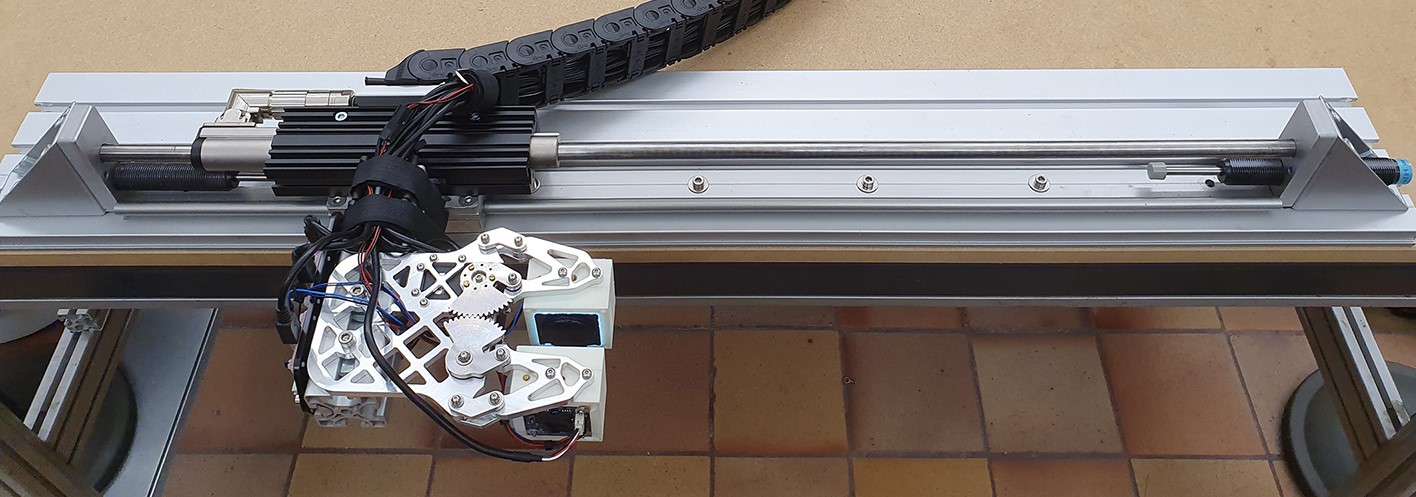}};

		\draw[fill=white, rounded corners=2pt, fill opacity=0.6, text opacity=1.0] (-0.5, -1.4) rectangle (1.0, -0.9) node[midway] {Sensor 2};
		\draw[fill=white, rounded corners=2pt, fill opacity=0.6, text opacity=1.0] (-0.4, 0.0) rectangle (1.1, -0.5) node[midway] {Sensor 1};
		\draw[fill=white, rounded corners=2pt, fill opacity=0.6, text opacity=1.0] (-3.2, -1.5) rectangle (-1.7, -1.0) node[midway] {Gripper};
		\draw[fill=white, rounded corners=2pt, fill opacity=0.6, text opacity=1.0] (-3.5, 1.5) rectangle (-1.3, 1.0) node[midway] {Linear motor};
	\end{tikzpicture}
	
	\caption{The robotic system presented in this paper consists of a parallel gripper comprising two tactile sensor, and a linear motor to which the gripper is mounted.}
	\vspace{-0.3cm}
	\label{fig:hardware}
\end{figure}
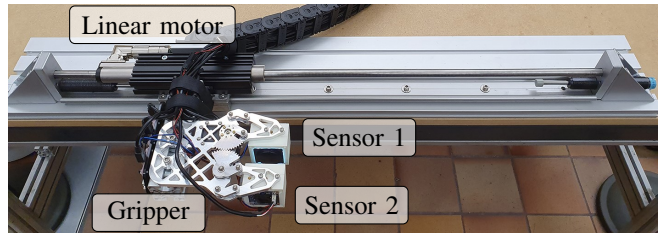

The robotic system considered in this paper consists of three main parts; a two-finger robotic gripper where each finger comprises a tactile sensor, a linear motor (stator/slider), to which the gripper is mounted, and finally, embedded computing systems that process the sensing data and send commands to the actuators. The linear motor and tactile gripper are pictured in \cref{fig:hardware}.

\subsection{Tactile Gripper}

\begin{figure}
	\centering
	\begin{subfigure}[b]{0.45\columnwidth}
		\centering
		\includegraphics[width=\textwidth]{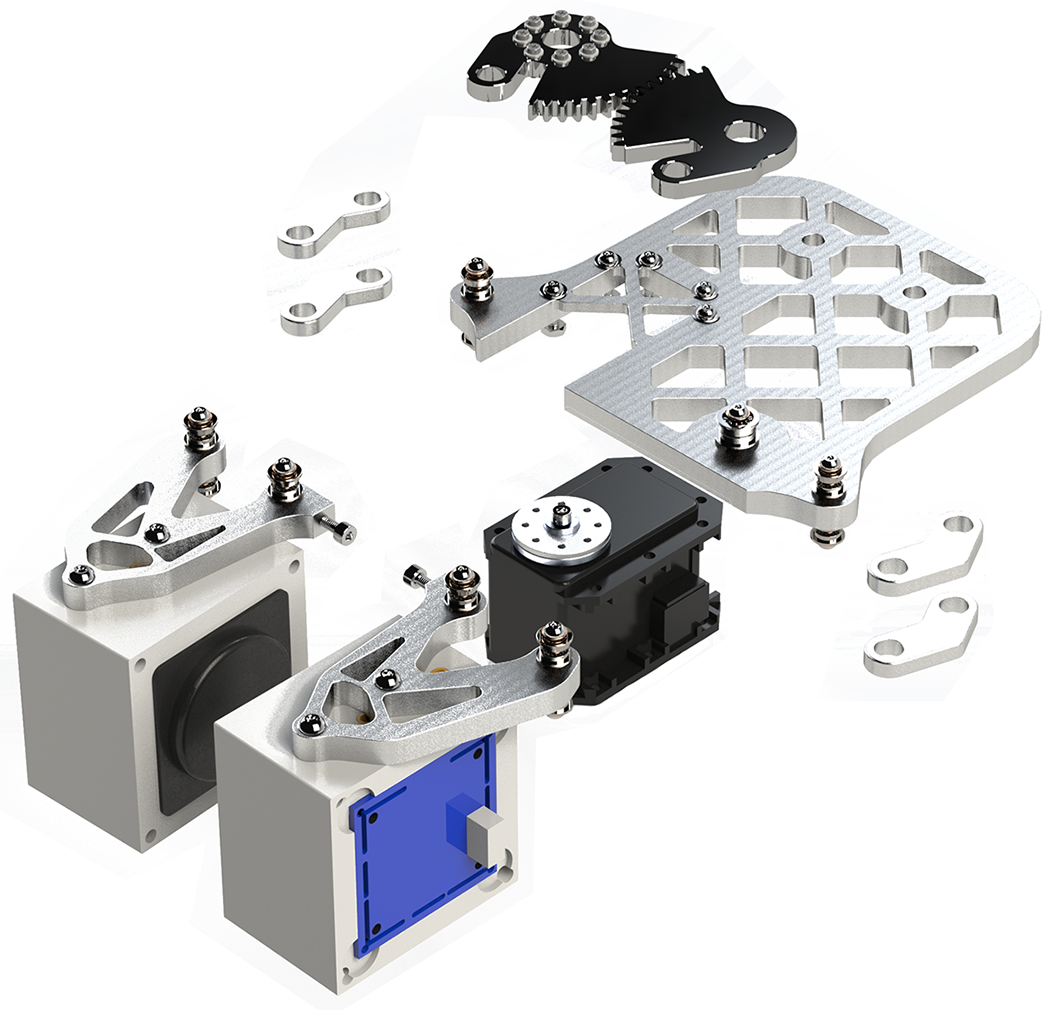}
		\caption{Parallel gripper}
		\label{fig:gripper_cad}
	\end{subfigure}
	\hfill
	\begin{subfigure}[b]{0.45\columnwidth}
		\centering
		\includegraphics[width=0.575\textwidth]{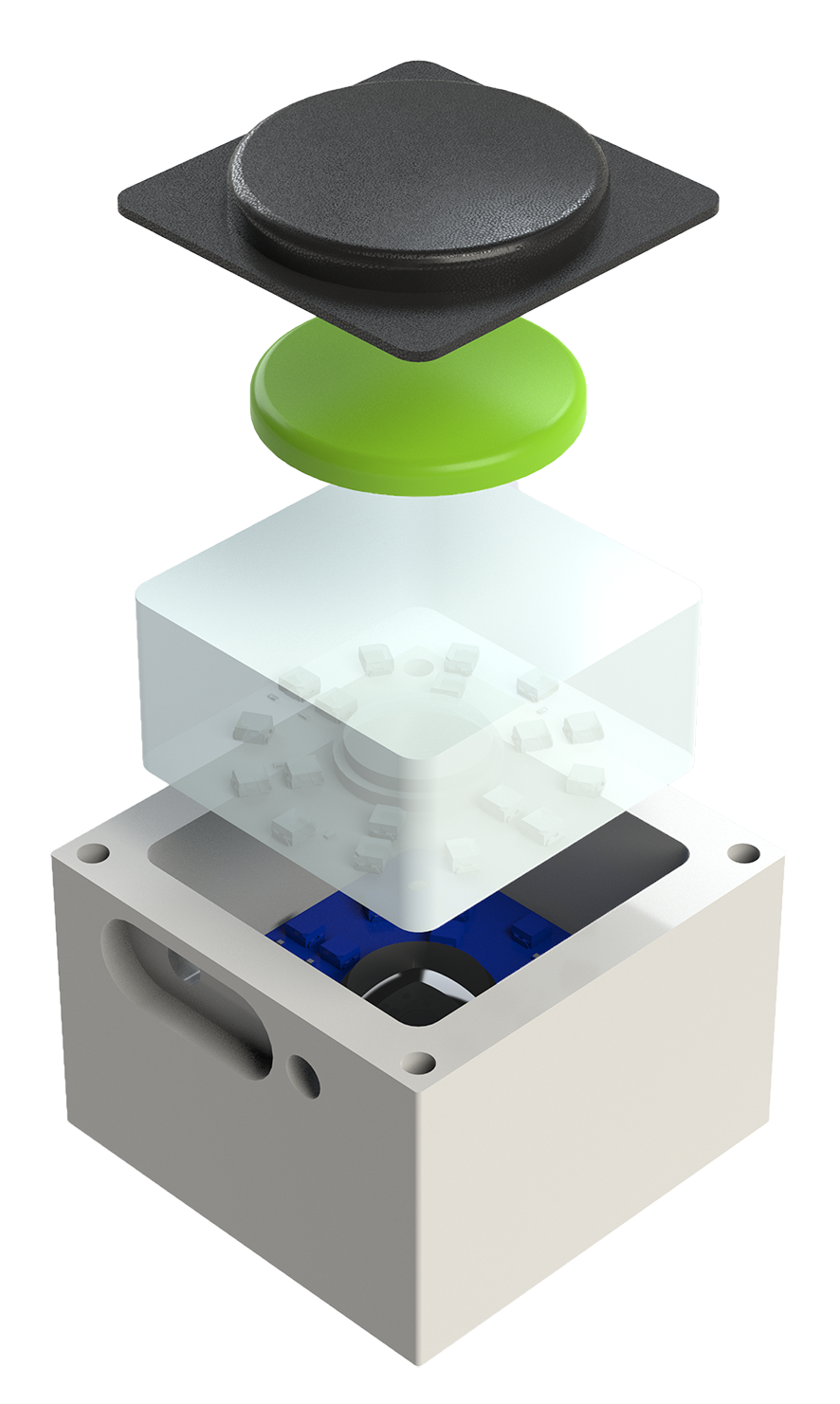}
		\caption{Tactile sensor}
		\label{fig:sensor_cad}
	\end{subfigure}
	\caption{This figure shows exploded views of the gripper (a) and the tacile sensor that acts as finger (b). A Dynamixel MX-28R controls the opening and closing of the two fingers, each of them equipped with a tactile sensor.}
	\vspace{-0.5cm}
	\label{fig:gripper_sensor}
\end{figure}

To enable the high-resolution gripping capability of the system, a custom 1-DoF parallel two-finger gripper was built in-house, see \cref{fig:gripper_cad}. A Dynamixel MX-28R servo motor is used to control the distance between the two fingers with a resolution of $\SI{0.06}{\milli\metre}$. %Further, it features an on-board PID position controller.  , specifically... symmetrical distance ...
Two tactile sensors, placed opposite each other, act as fingers for the gripper.
The sensing principle employed in this paper is based on \cite{sferrazza2019design}. Three soft silicone layers are poured on top of an RGB fisheye camera (ELP USBFHD06H), surrounded by LEDs. The base layer (ELASTOSIL
\textregistered{} RT 601 RTV-2, mixing ratio 7:1,
shore hardness 45A) is stiff and transparent, and serves as a spacer. The middle layer (ELASTOSIL
\textregistered
RT 601 RTV-2, ratio 25:1, shore hardness 10A) is soft and transparent, and embeds a spread of randomly distributed fluorescent green particles. A black top layer (made of the same material as the middle layer) completes the sensor and shields it from external light disturbances. 
The soft sensor's surface is slightly curved to provide a more anatomical grasping surface.
An exploded view
of the sensor layers is shown in \cref{fig:sensor_cad}.

\subsection{Linear Motor}
In order to achieve the translational motion of the gripper, a linear motor comprising a \emph{stator} and a \emph{slider} is employed. The stator (LinMot P01-23x160H-HP-R) contains the motor windings, bearings for the slider, position capture sensors and a microprocessor, and is thus able to generate motion with respect to the slider. The slider (LinMot PL01-12x850/810-HP) is a stainless steel tube and is fixed to a table so that the stator is the only moving part. The gripper is then mounted to the stator through the use of a motor flange (LinMot PF02-23x120). A motor drive (LinMot C1100-GP-XC-0S-000) controls the motion of the stator.
%The final setup has an effective stroke of $\SI{0.53}{\metre}$. Accelerations of up to $\SI{40}{\metre\per\second\squared}$ and velocities up to $\SI{3.5}{\metre\per\second}$ are successfully tracked within the entire range of the stroke.

\subsection{Embedded Systems}
Two embedded devices are used to control the system. First, a Raspberry Pi (RPi) runs two low-level controllers, one for the linear motor and one for the gripper. The linear motor controller tracks commanded acceleration setpoints, while the gripper controller tracks the distance between the two fingers.
Second, an NVIDIA Jetson TX2, a compact embedded device with a built-in GPU, obtains the camera images from the tactile sensor at $\SI{60}{\hertz}$, pre-processes the images, and infers the force distribution. Note that this pipeline is only executed for sensor 1 (see \cref{fig:hardware}). This is motivated by the fact that due to the planar nature of the system, the forces acting on sensor 2 can be assumed to be symmetrical to those acting on sensor 1. Furthermore, this reduces the computational complexity of the pipeline.

The Jetson also receives the current actuator states from the RPi. Control actions are then inferred using the proposed control policy and are communicated to the low-level controllers on the RPi that execute the commands.

\section{METHOD}
\label{sec:method}

The proposed method can be divided into three different parts. First, the vision-based tactile sensor estimates the force distribution acting on its surface from its camera images. Second, a simulator for the dynamics of a pole and the given robotic system is developed. Third, a tactile feedback control policy for the swing-up manipulation is learned in the simulation using reinforcement learning.

\subsection{Vision-Based Tactile Sensing}
\label{sec:sens}

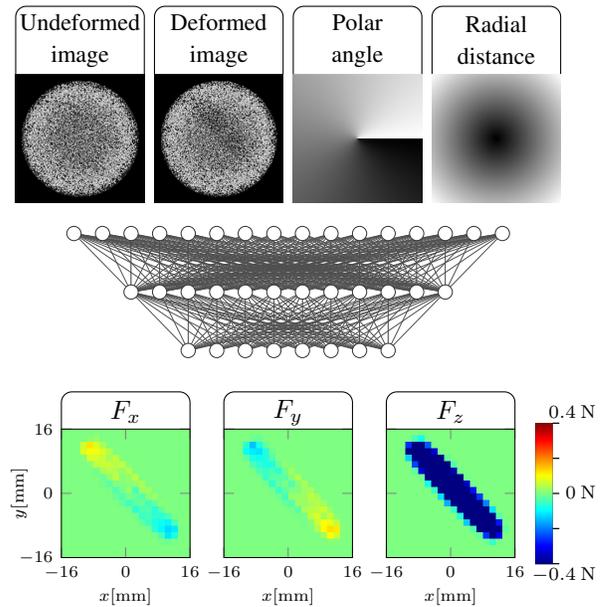
\begin{figure}%[!ht]
	\centering
	\begin{tikzpicture}[scale=0.9]
		
		%\node[rotate=90] at (-1.2, 0.45) {Input channels};
		
		\node[inner sep=0pt] at (0,0)
		{\includegraphics[width=1.71cm]{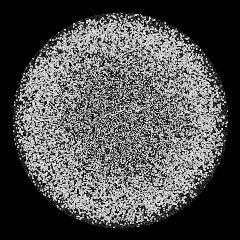}};
		
		\node[inner sep=0pt] at (2.05,0)
		{\includegraphics[width=1.71cm]{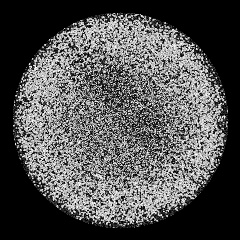}};
		
		\node[inner sep=0pt] at (4.1, 0)
		{\includegraphics[width=1.71cm, angle=180,origin=c]{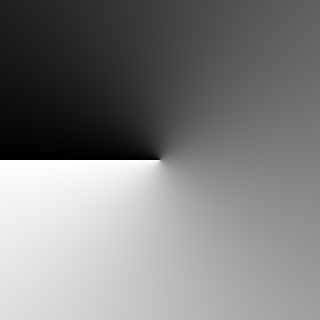}};
		
		\node[inner sep=0pt] at (6.15, 0)
		{\includegraphics[width=1.71cm]{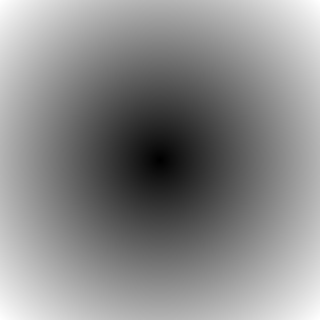}};

		\draw[fill=white] {(-0.95, 0.95) {[rounded corners] 
				-- (-0.95, 1.95) 
				-- (0.95, 1.95)}
			-- (0.95, 0.95)
		};
		\node[text centered, text width=1.7cm] at (0, 1.45) {\small Undeformed image};

		\draw[fill=white] {(-0.95+2.05, 0.95) {[rounded corners] 
				-- (-0.95+2.05, 1.95) 
				-- (0.95+2.05, 1.95)}
			-- (0.95+2.05, 0.95)
		};
		\node[text centered, text width=1.9cm] at (0+2.05, 1.45) {\small Deformed image};
		
		\draw[fill=white] {(-0.95+2*2.05, 0.95) {[rounded corners] 
				-- (-0.95+2*2.05, 1.95) 
				-- (0.95+2*2.05, 1.95)}
			-- (0.95+2*2.05, 0.95)
		};
		\node[text centered, text width=1.3cm] at (0+2*2.05, 1.45) {\small Polar angle};
		
		\draw[fill=white] {(-0.95+3*2.05, 0.95) {[rounded corners] 
				-- (-0.95+3*2.05, 1.95) 
				-- (0.95+3*2.05, 1.95)}
			-- (0.95+3*2.05, 0.95)
		};
		\node[text centered, text width=1.5cm] at (0+3*2.05, 1.45) {\small Radial distance };
		
		%		\draw[fill=white, rounded corners] (-0.2, -0.25) rectangle (3.2, 0.2) node[midway] {Deformed image};
		%		\draw[fill=white, rounded corners] (1.3, -1.75) rectangle (4.7, -1.3) node[midway] {Polar angle};
		%		\draw[fill=white, rounded corners] (2.8, -3.25) rectangle (6.2, -2.8) node[midway] {Radial distance};
		
		%		\node at (2.25, -7) {(a) Input channels};
		%		\node at (10.0, -7) {(b) Output channels};
		
		% Network
		\node at (3.075, -2.3) {\includegraphics[width=6cm]{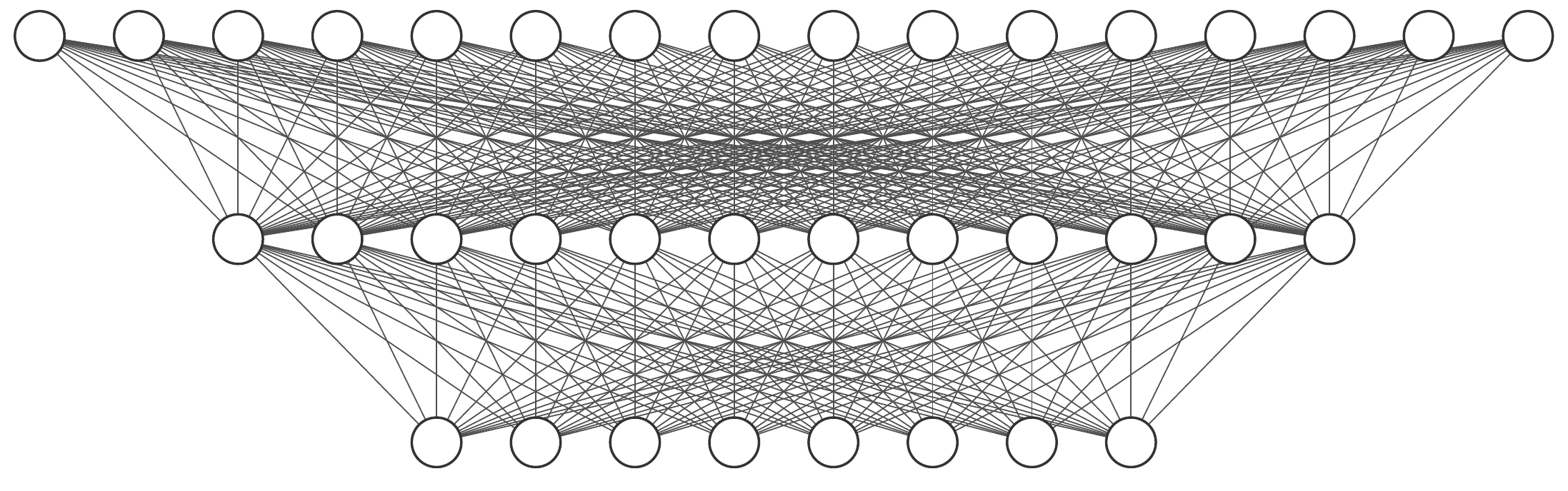}};

		%		\node[inner sep=0pt] at (3.075-2.5, -5.25)
		%		{\includegraphics[width=1.9cm, angle=180]{images/forces_6_0.jpg}};
		%		\node[inner sep=0pt] at (3.075, -5.25)
		%		{\includegraphics[width=1.9cm, angle=180]{images/forces_6_1.jpg}};
		%		\node[inner sep=0pt] at (3.075+2.5, -5.25)
		%		{\includegraphics[width=1.9cm, angle=180]{images/forces_6_2.jpg}};

		\draw[fill=white] {(-0.95+3.075-2.4, -4.3) {[rounded corners] 
				-- (-0.95+3.075-2.4, -3.75) 
				-- (0.95+3.075-2.4, -3.75)}
			-- (0.95+3.075-2.4, -4.3)
		};
		\node[text centered, text width=1.9cm] at (3.075-2.4, -4.05) {$F_x$};
		\draw[fill=white] {(-0.95+3.075, -4.3) {[rounded corners] 
				-- (-0.95+3.075, -3.75) 
				-- (0.95+3.075, -3.75)}
			-- (0.95+3.075, -4.3)
		};
		\node[text centered, text width=1.9cm] at (3.075, -4.05) {$F_y$};
		\draw[fill=white] {(-0.95+3.075+2.4, -4.3) {[rounded corners] 
				-- (-0.95+3.075+2.4, -3.75) 
				-- (0.95+3.075+2.4, -3.75)}
			-- (0.95+3.075+2.4, -4.3)
		};
		\node[text centered, text width=1.9cm] at (3.075+2.4, -4.05) {$F_z$};

		\begin{axis}[
			xtick distance=48,
			ytick distance=48,
			xticklabels={,$\scriptstyle -16$, $\scriptstyle 0$, $\scriptstyle 16$},
			yticklabels={,$\scriptstyle -16$, $\scriptstyle 0$, $\scriptstyle 16$},
			x label style={at={(axis description cs:0.5,0.1)},anchor=north},
			xlabel={$\scriptstyle x[\SI{}{\milli\metre}]$},
			y label style={at={(axis description cs:0.45,0.5)},anchor=south},
			ylabel={$\scriptstyle y[\SI{}{\milli\metre}]$},
			anchor=center,
			%			xtick align=outside,
			width=1.9cm,
			height=1.9cm,
			at={(0.675cm, -5.25cm)},
			scale only axis,
			enlargelimits=false,
			axis on top]
			\addplot graphics[xmin=0,xmax=96,ymin=0,ymax=96] {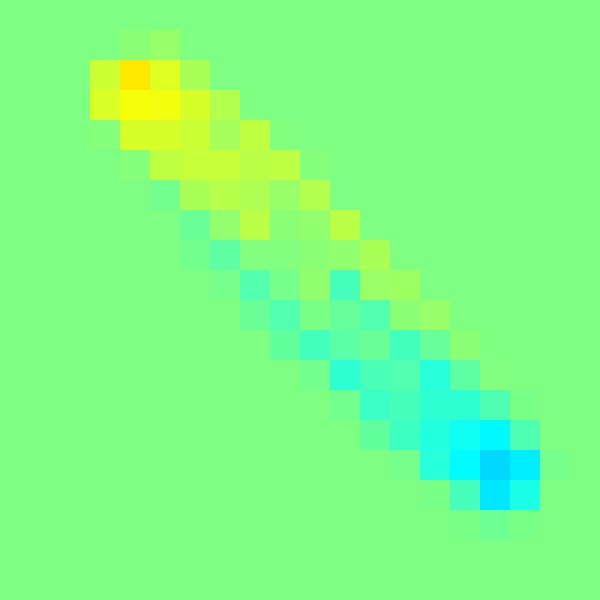};
		\end{axis}
		\begin{axis}[
			xtick distance=48,
			ytick distance=48,
			xticklabels={,$\scriptstyle -16$, $\scriptstyle 0$, $\scriptstyle 16$},
			yticklabels={},%$\scriptstyle -16.5$, $\scriptstyle 0.0$, $\scriptstyle 16.5$},
			x label style={at={(axis description cs:0.5,0.1)},anchor=north},
			xlabel={$\scriptstyle x[\SI{}{\milli\metre}]$},
			anchor=center,
			%			xtick align=outside,
			width=1.9cm,
			height=1.9cm,
			at={(3.075cm, -5.25cm)},
			scale only axis,
			enlargelimits=false,
			axis on top]
			\addplot graphics[xmin=0,xmax=96,ymin=0,ymax=96] {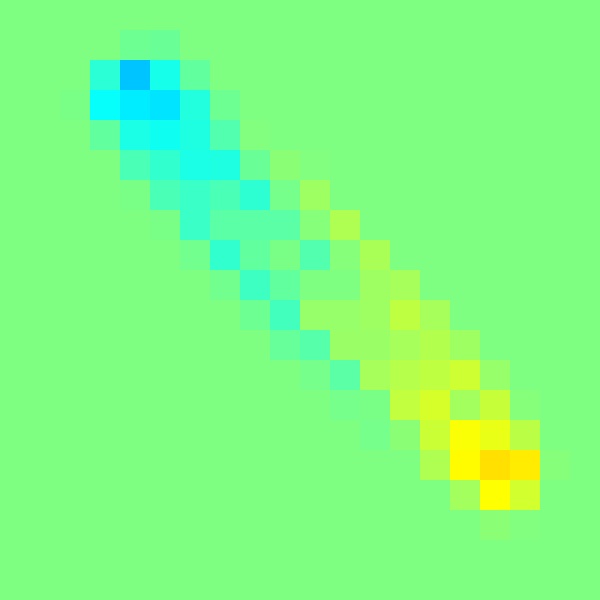};
		\end{axis}
		\begin{axis}[
			xtick distance=48,
			ytick distance=48,
			xticklabels={,$\scriptstyle -16$, $\scriptstyle 0$, $\scriptstyle 16$},
			yticklabels={},%$\scriptstyle -16.5$, $\scriptstyle 0.0$, $\scriptstyle 16.5$},
			x label style={at={(axis description cs:0.5,0.1)},anchor=north},
			xlabel={$\scriptstyle x[\SI{}{\milli\metre}]$},
			anchor=center,
			%			xtick align=outside,
			width=1.9cm,
			height=1.9cm,
			at={(5.475cm, -5.25cm)},
			scale only axis,
			enlargelimits=false,
			axis on top]
			\addplot graphics[xmin=0,xmax=96,ymin=0,ymax=96] {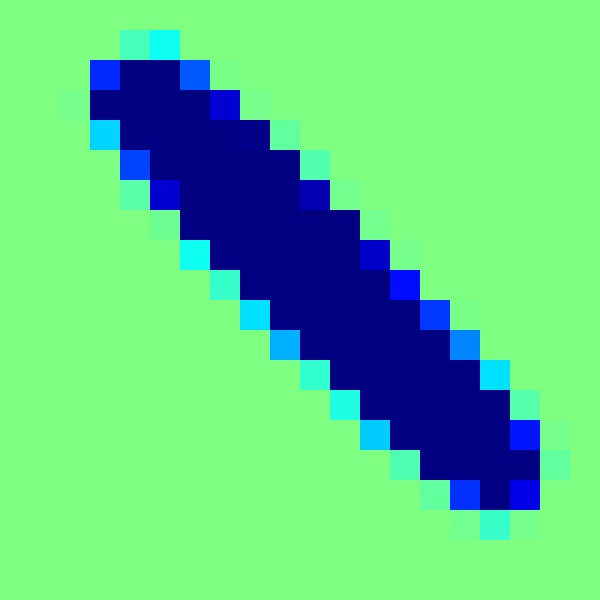};
		\end{axis}
	
		\node[inner sep=0] at (6.9, -5.25) {\includegraphics[height=1.9cm]{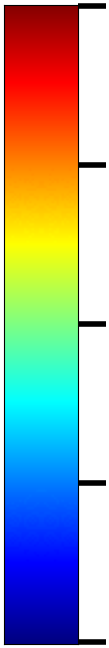}};
		\node at (6.95+0.2, -5.25-1.2) {$\scriptsize \SI{-0.4}{\newton}$};
		\node at (6.95+0.2, -5.25+1.2) {$\scriptsize \phantom{-}\SI{0.4}{\newton}$};
		\node at (6.95+0.2, -5.25) {$\scriptsize \phantom{-0.}\SI{0}{\newton}$};

		%		\draw[fill=white, rounded corners] (6.8, 0.95) rectangle (10.2, 0.5) node[midway] {$F_x$};
		%		\draw[fill=white, rounded corners] (8.3, -0.55) rectangle (11.7, -1.0) node[midway] {$F_y$};
		%		\draw[fill=white, rounded corners] (9.8, -2.05) rectangle (13.2, -2.5) node[midway] {$F_z$};

	\end{tikzpicture}
	\caption{The features and labels of our supervised learning task. The features are the images of the particles in the undeformed and in the deformed states. Two additional channels provide the polar coordinates of the pixels. For the labels, $F_x, F_y$ and $F_z$ denote the $x, y$ and $z$ components of the discretized force distribution. The force distributions shown in the figure were collected with a pole pivoting on the sensor (as in the experiments discussed in \mbox{Section \ref{sec:results}}). They show how the shape and pose of the pole manifest themselves correctly in the force distribution readings. Additionally, the lateral forces ($F_x$ and $F_y$) on either side of the center of rotation point in opposite directions, which can be deducted from their change in color.}
	\label{fig:shuffle}
	\vspace{-0.5cm}
\end{figure}

The tactile sensor employed in this paper follows the same sensing principle as introduced in \cite{sferrazza2019design}. When the soft sensing surface is subject to force, the
material deforms and displaces the particles tracked by
the camera. This motion generates different patterns in the
images. The material deformation at any point in time can thus be described by two camera images, one where no loads are applied and the material is at rest, and another at the current deformed state.

In \cite{softrob}, a method to generate such images in simulation is presented to train a supervised learning architecture that aims to accurately estimate the real-world 3D contact force distribution. The same approach to generate training data is employed here, using finite-element simulations of the sensor surface under various contact conditions, where hyperelastic material models for the sensor's soft materials are employed. The details of this procedure can be found in \cite{softrob}. In addition to the two mentioned images per datapoint, the polar coordinates of each pixel are encoded here as two additional image channels. Explicitly incorporating such spatial location features has previously been shown to significantly improve accuracy where the location of image features is relevant for the task at hand \cite{ghafoorian2017location}.
Using a fully convolutional neural network based on ShuffleNet V2 \cite{ma2018shufflenet}, the resulting four-channel image is then mapped to accurate contact force distribution labels (see \cref{fig:shuffle}), with ground truth also obtained from finite element simulations \cite{sferrazza2019ground}.  %The network design is based on ShuffleNet V2 \cite{ma2018shufflenet}. Using the same naming convention as in \cite{ma2018shufflenet}, the following modifications are applied to the original architecture: Stage2 features 64 output channels, and Stage3 128 output channels. Additionally, all the layers after Stage3 are discarded and replaced with a single 1$\times$1 convolutional layer that reduces the final number of channels to three (one for each component of the force distribution).

\begin{figure}
	\centering
	\begin{subfigure}[b]{0.31\columnwidth}
		\centering
		\includegraphics[width=\textwidth]{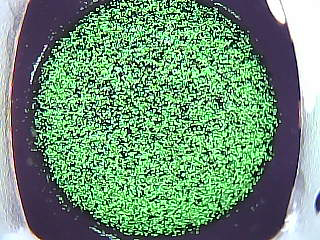}
		\caption{Original image}
		\label{fig:y equals x}
	\end{subfigure}
	\hspace{0.05cm}
	\begin{subfigure}[b]{0.31\columnwidth}
		\centering
		\includegraphics[width=\textwidth]{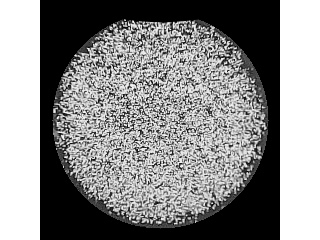}
		\caption{Remapped image}
		\label{fig:y equals x}
	\end{subfigure}
	\hspace{0.05cm}
	\begin{subfigure}[b]{0.31\columnwidth}
		\centering
		\includegraphics[width=\textwidth]{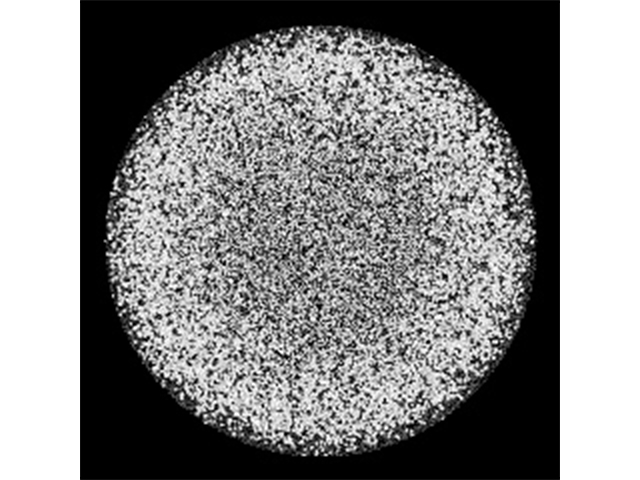}
		\caption{Simulated image}
		\label{fig:y equals x}
	\end{subfigure}
    \caption{Camera images from the real-world sensor are preprocessed converting the original image (a) to gray-scale, and remapping the image using a calibrated camera model (b). The last image (c) corresponds to a simulated image that is used for training and is provided for comparison.}
\label{fig:remap}
\vspace{-0.3cm}
\end{figure}

On the real-world sensor, camera images are preprocessed to match those of the simulated training dataset as described in \cite{softrob} and \cite{sferrazza2020learning}. Specifically, images are converted to gray-scale and remapped using the real-world camera model (obtained via a state-of-the-art calibration technique \cite{scaramuzza}) to images of the same scene as if they were taken in the simulated world. A circular mask is then applied to remove any irrelevant image information. The results of this preprocessing procedure are illustrated in \cref{fig:remap}. On the given hardware, the force distribution for a given preprocessed camera image can be inferred in real-time in $\SI{2.5}{\milli\second}$.

\subsection{Tactile Simulation}
\label{sec:sim}

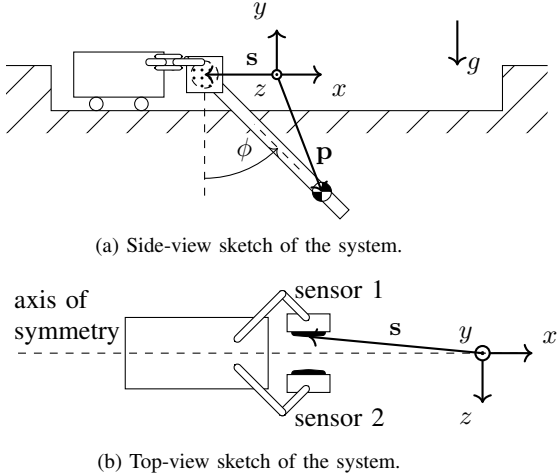
\begin{figure}
	\centering
	
	\begin{subfigure}[b]{0.75\columnwidth}
		\centering
		\begin{tikzpicture}[scale=0.6]

			% Motor guide
			\draw (-6, 1) -- (-5, 1) -- (-5, 0) -- (5, 0) -- (5, 1) -- (6, 1);
			\path[pattern=south west lines] (-6, -0.5) -- (-6, 1) -- (-5, 1) -- (-5, 0) -- (5, 0) -- (5, 1) -- (6, 1) -- (6, -0.5) -- cycle;
			
			% Pole
			\draw[fill=white] (-1.2-0.2, 0.4)  -- (1.5-0.1, -2.3-0.1) --  (1.5+0.1, -2.3+0.1) -- (-1.2, 0.4+0.2);
			\node at(1.0, -1.8) {\small\centerofmass};
			
			% Pole dashed angle
			\draw[dashed] (0.5, -1.3) -- (-0.5, -0.3);
			
			% Axis
			\draw[dashed] (-1.6, 0.4) -- (-1.6, -2.0);
			\draw[->] (-1.6, -1.5) arc (-90:-45:2.3) node[above, midway] {$\phi$};
			
			% Stator
			\draw (-2.5, 0.3) rectangle (-4.5, 1.3);
			\draw (-3, 0.15) circle (0.15);
			\draw (-4, 0.15) circle (0.15);
			
			% Sensor
			\draw (-2, 0.4) rectangle (-1.2,1.2);  % Sensor
			\draw[dashed, pattern=dots] (-1.6, 0.8) circle (0.3);  % Sensor
			
			% Gripper
			\draw[fill=white] (-2.0, 1.15) circle (0.05);
			\draw[fill=white] (-2.3, 0.95) circle (0.05);
			\draw[rounded corners=0.04cm, fill=white] (-2.7, 1.15) rectangle (-2.1, 1.25);
			\draw[rounded corners=0.04cm, fill=white] (-2.7, 0.9) rectangle (-2.1, 1.0);
			\draw[rounded corners=0.04cm, fill=white] (-2.3, 1.0) rectangle (-1.6, 1.15);
			\draw[rounded corners=0.04cm, fill=white] (-2.9, 1.0) rectangle (-2.45, 1.15);

			% Stroke length
			%\draw[<->] (-5, -1) -- (5, -1);

			% Position vector
			\draw[thick, ->] (0.0, 0.8) -- (-1.6, 0.8) node[anchor=south west] {$\hspace{12pt}\mathbf{s}$};
			
			\draw[thick, ->] (0.0, 0.8) -- (1.0, -1.8);
			\node at (0.7, -1.0) {$\hspace{12pt}\mathbf{p}$};

			% Axis
			\draw[thick,->] (0,0.8) -- (1.0,0.8) node[anchor=north west] {$x$};
			\draw[thick,->] (0,0.8) -- (0,1.8) node[anchor=south east] {$y$};
			\draw[thick, fill=white] (0.0,0.8) circle (0.1) node[anchor=north east] {$z$};
			\draw[fill=black] (0.0, 0.8) circle (0.02);
			
			\draw[thick, ->] (4.0, 2.0) -- (4.0, 1.0) node[right] {$g$};

		\end{tikzpicture}
		\caption{Side-view sketch of the system.}
		\label{fig:Ng1} 
	\end{subfigure}
	
	\vspace{1ex}
	
	\begin{subfigure}[b]{0.75\columnwidth}
		\centering
		\begin{tikzpicture}[scale=0.95]
			
			% Axis
			\draw[thick,->] (2,0.0) -- (2.7,0.0) node[anchor=south west] {$x$};
			\draw[thick,->] (2,0.0) -- (2,-0.7) node[anchor=north east] {$z$};
			\draw[thick, fill=white] (2.0,0.0) circle (0.1) node[anchor=south east] {$y$};
			\draw[fill=black] (2.0, 0.0) circle (0.02);
			
			% Cart
			\draw (-3.0, 0.5) rectangle (-1.0, -0.5);
			
			\draw[dashed] (-4.5, 0) -- (2.5, 0);
			\node[text width=2cm] at (-3.5, 0.5) {axis of symmetry};

			%Sensors
			\draw[rounded corners=1pt, fill=black] (-0.2, 0.3) -- (-0.2, 0.25) -- (-0.435, 0.23) -- (-0.67, 0.25) -- (-0.67, 0.3);
			\draw[rounded corners=1pt, fill=black] (-0.2, -0.3) -- (-0.2, -0.25) -- (-0.435, -0.23) -- (-0.67, -0.25) -- (-0.67, -0.3); 
			
			\draw[fill=white] (-0.435-0.3, 0.55) rectangle (-0.435+0.3, 0.3);
			\draw[fill=white] (-0.435-0.3, -0.55) rectangle (-0.435+0.3, -0.3);
			
			% Gripper
			\draw[fill=white,rounded corners=1.5pt] (-0.8-0.07, 0.9-0.07) -- (-0.8, 0.9) -- (-0.4, 0.5) -- (-0.4-0.07, 0.5-0.07) -- cycle;
			
			\draw[fill=white,rounded corners=1.5pt] (-0.8-0.07, -0.9+0.07) -- (-0.8, -0.9) -- (-0.4, -0.5) -- (-0.4-0.07, -0.5+0.07) -- cycle;
			
			\draw[fill=white,rounded corners=1.5pt] (-1.5+0.07, -0.2+0.07) -- (-1.5, -0.2) -- (-0.8, -0.9) -- (-0.8+0.07, -0.9+0.07) -- cycle;
			
			\draw[fill=white, rounded corners=1.5pt] (-1.5+0.07, 0.2-0.07) -- (-1.5, 0.2) -- (-0.8, 0.9) -- (-0.8+0.07, 0.9-0.07) -- cycle;
			
			% Position vector
			\draw[thick, ->] (2, 0) -- (-0.435, 0.23) node[above, midway] {$\mathbf{s}$};
			
			% sensor labels
			\node at (0.0, 0.9) {sensor 1};
			\node at (0.0, -0.9) {sensor 2};

		\end{tikzpicture}
		\caption{Top-view sketch of the system.}
		\label{fig:Ng1} 
	\end{subfigure}
	\caption{The system can be described as a cart-pole augmented with a parallel gripper that features the two sensors.} %This geometric sketch describes the coordinate system as well as the position vector of the pole $\mathbf{p}$ and the sensor $\mathbf{s}$, and the orientation $\phi$ of the pole. (a) provides a view from the side, while (b) shows the system from above.}
	\label{fig:sim}
	\vspace{-0.5cm}
\end{figure}

%The non-linear finite element analysis approach using hyperelastic material models (as done for the supervised training of the sensor) is computationally expensive and not suitable for the use in a reinforcement learning setting, where a large number of simulated episodes may be necessary to reach satisfactory performance.

%Here, two .. to build a fast simulator. 

In order to achieve a fast simulator, essential for training reinforcement learning algorithms in a reasonable amount of time, a few model simplifications are introduced here.
First, the material of the sensor is assumed to be linearly elastic. %, allowing to use finite elements... 
%Second, the planar nature of the problem is exploited \textcolor{red}{doesn't this decoupling also hold for non-planar motions?} by decoupling the forces acting on the sensor surface into two components: the forces arising due to the material deformation in the $z$ direction, and the lateral friction forces resulting from the relative motion of the pole  with respect to the sensor. A real-time finite element approach is then employed to compute the two mentioned force components.
Second, the forces acting on the sensor surface are decoupled into two components: the forces arising due to the material deformation in the $z$ direction, and the lateral friction forces resulting from the relative motion of the pole  with respect to the sensor. A real-time finite element approach is then employed to compute the two mentioned force components.

%Further, .. discretized dynamics..

%Semi-implicit velocity-stepping methods \cite{hwangbo2018per, todorov2012mujoco}

\subsubsection{Problem Statement}

A sketch of the system considered in this work is shown in \cref{fig:sim}, where a single coordinate system is defined.
\hl{The $x$-axis is aligned to the moving axis of the linear motor, denoted in the following as the cart. The $y$-axis points in the opposite direction to gravity, i.e. upwards. Finally, the $z$-axis is chosen such that all the points on \mbox{sensor 1} exhibit a negative $z$-coordinate. The reference position  $\mathbf{s}=(x_\text{s},\ y_\text{s},\ z_\text{s})$ is chosen with the point on the curved surface of sensor 1 that is closest to the $x$-$y$ plane (at rest).} %The position of sensor 2 
While there are two tactile sensors, their positions are symmetrical about the $x$-$y$ plane. Hence, it suffices to only consider the position of a single sensor. Next, the orientation of the pole is defined by the angle $\phi$. Lastly, $\mathbf{p}=(x_\text{p},\ y_\text{p},\ z_\text{p})$ denotes the position of the center of mass of the pole.
Note that the sensor is fixed in the $y$-direction ($y_\text{s}=0$) and the pole is fixed in the $z$-direction ($z_\text{p}=0$).
These definitions are illustrated in \cref{fig:sim}.
The state vector $\mathbf{x}$ is then defined as
\begin{align}
	\mathbf{x}&=( x_\text{s},\ \dot{x}_\text{s},\ z_\text{s},\ x_\text{p},\ \dot{x}_\text{p},\ y_\text{p},\ \dot{y}_\text{p},\ \phi,\ \dot{\phi})
\end{align}
Since only the static behavior of the sensor material is analyzed, $\dot{z}_\text{s}$ is not considered.

The inputs to the system are the cart acceleration and the increment in $z_s$ between two subsequent timesteps:
\begin{align}
	\mathbf{u}&=( \ddot{x}_\text{s},\ \Delta z_\text{s} )\quad .
\end{align}
Note that on the real system, the servo commands are mapped to $\Delta z_\text{s}$ with a linear mapping identified from data.

Next, the pole is characterized as a rigid cylinder. Its radius is given by $r_\text{p}$, the mass by $m_\text{p}$ and its moment of inertia about the center of mass and along the $z$-axis by $I_\text{p}$. The length of the pole above its center of mass is given by $l_{\text{p},\text{u}}$ and the length below the center of mass by $l_{\text{p},\text{l}}$.
%Since dynamic effects of the sensor (e.g. restitution) are not considered here, 

Given these definitions, the goal is to model the state evolution over time, i.e., 
	$\mathbf{x}(k+1) = f(\mathbf{x}(k), \mathbf{u}(k))$,
where $k$ is the discrete time index, and $f$ describes a functional dependency. In the following, the time index $(k)$ will be omitted, and variables at time $(k+1)$ will be denoted by a $+$ superscript, e.g. $x^+$.
%The above equation  , e.g. $\mathbf{x}^+=f(\mathbf{x}, \mathbf{u})$

\subsubsection{Equations of Motion}
%As the motion of the cart, and thus the sensor can be directly controlled. 
The pole is modeled as a free-body constrained to move in the $x$-$y$ plane, meaning that its motion is governed by the force $F_\text{p}=(
    F_{\text{p},x},\ F_{\text{p},y},\  0)$ and torque $T_\text{p}=(
    0,\ 0,\ T_{\text{p},z}
)$ acting on its center of mass.
Using a semi-implicit integration scheme (\cite{hwangbo2018per, todorov2012mujoco}) the equations of motion are then given by
\begin{align}
	\dot{x}_\text{s}^+ &= \dot{x}_\text{s} + \Delta t\ \ddot{x}_\text{s}  &
	x_\text{s}^+ &= x_\text{s} + \Delta t\ \dot{x}_\text{s}^+\label{eq:eomfirst}\\  % TODO
	\dot{x}_\text{p}^+ &= \dot{x}_\text{p} + \Delta t\ \frac{F_{\text{p},x}}{m_\text{p}} &
	x_\text{p}^+ &= x_\text{p} + \Delta t\ \dot{x}_\text{p}^+\\
	\dot{y}_\text{p}^+ &= \dot{y}_\text{p} + \Delta t\ \frac{F_{\text{p},y}}{m_\text{p}}&
	y_{\text{p}}^+ &= y_{\text{p}} + \Delta t\ \dot{y}_\text{p}^+\\
	\dot{\phi}^+ &= \dot{\phi} + \Delta t\ \frac{T_{\text{p},z}}{I_\text{p}}&
	\phi^+ &= \phi + \Delta t\ \dot{\phi}^+\\
	& & z_\text{s}^+ &= z_\text{s}+\Delta z_\text{s}\label{eq:eomlast} % \text{NEED TO DISCUSS!} 
\end{align}
%Note that this integration scheme is key to achieving physically consistent behaviours {\color{red}of contact dynamics}.
In the following, the derivation of $F_\text{p}$ and $T_\text{p}$ is presented.

%\subsubsection{Contact Resolution}

%Since the pole is in contact with the sensors over continuous areas..finite elements.. %As mentioned previously, the finite element method can be used to discretize deformable bodies into finite elements/nodes which form a so-called mesh. 
\hl{Both sensors are discretized using an identical mesh of $N=576$ finite elements (nodes).}
Hereafter, all quantities introduced will refer to sensor 1, where the corresponding counterparts of sensor 2 are clear from the symmetrical context and are denoted using a tilde, i.e. $\tilde{\cdot}$. Then, for node $i$ of the mesh, let $(x_{i}, y_{i}, z_{i})$ be its coordinates, and $F_i$ the force acting on the node. Each node $i$ in contact with the pole leads to a planar reaction force 
\begin{align}
\label{eq:equiv}
    F_{\text{p},i} = -(F_{i}+\tilde{F}_{i}) \implies F_{\text{p},i,x:y}=-2F_{i,x:y}  %\quad \forall i \in \mathcal{C} 
\end{align}
where the implication follows from symmetry, with the $x\!:\!y$ subscript denoting the stacked $x$ and $y$ components of the three-dimensional vector.
%Note that.. due to .. $F_\text{p}$, known from $F_i$ and vice-versa..
Next, the gravitational force acting on the pole is denoted by
$F_\text{g}=\begin{pmatrix} 0,\ -m_\text{p} g,\ 0 \end{pmatrix}$,
where \mbox{$g=\SI{9.81}{\metre\per\second\squared}$}. Defining $\mathbf{r}_{i}:=\begin{pmatrix} x_{i}-x_{\text{p}},& y_{i}-y_{\text{p}}, & 0 \end{pmatrix}$, the total force and torque acting on the pole are then
\begin{align}
    F_{\text{p}} &= F_{g} + \sum_{i\in\mathcal{S}} {F}_{\text{p},i} \ , &
    T_\text{p} &= \sum_{i\in\mathcal{S}} \mathbf{r}_{i}\times {F}_{\text{p},i}. \label{eq:t_tot}
\end{align}
\hl{where $\mathcal{S}$ is the set of all nodes on the surface of the sensor.}
%\Cref{fig:fem} displays the sets $\mathcal{C}_k$, $\mathcal{F}_k$ and $\mathcal{N}_k$.

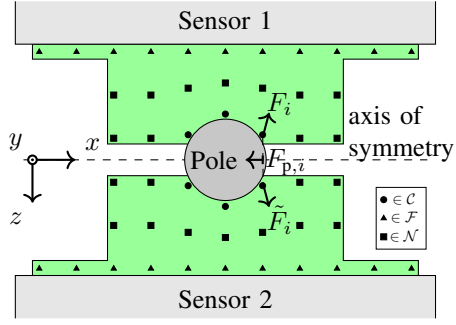
\begin{figure}
	\centering
	\begin{tikzpicture}[scale=0.57]

		% Sensor outlines
		\draw[fill=green!40!white] (5.21-0.88, 2.33) -- (5.21-2.75, 2.33) -- (5.21-2.75, 0.35) -- (5.21-4.5, 0.35) -- (5.21-4.5, 0.0) -- (5.21+4.5, 0.0) -- (5.21+4.5, 0.35) -- (5.21+2.75, 0.35) -- (5.21+2.75, 2.33) -- (5.21+0.88, 2.33);
		
		\draw[fill=green!40!white] (5.21-0.88, 3.07) -- (5.21-2.75, 3.07) -- (5.21-2.75, 5.05) -- (5.21-4.5, 5.05) -- (5.21-4.5, 5.4) -- (5.21+4.5, 5.4) -- (5.21+4.5, 5.05) -- (5.21+2.75, 5.05) -- (5.21+2.75, 3.07) -- (5.21+0.88, 3.07);
		
		\draw[fill=black!10!white] (5.21-4.9, -1.0) -- (5.21-4.9, 0.0) -- (5.21+4.9, 0.0) -- (5.21+4.9, -1.0);
		\draw[fill=black!10!white] (5.21-4.9, 5.4+1.0) -- (5.21-4.9, 5.4) -- (5.21+4.9, 5.4) -- (5.21+4.9, 5.4+1.0);
		
		% Axis
		\node[text width=1.5cm] at (9.4, 3.3) {axis of symmetry};
		\draw[dashed] (0.5, 2.7) -- (9.92, 2.7);
		
		% Pole
		\draw[fill=black!22!white] (5.21, 2.7) circle (0.95) node[xshift=-1.5mm] {Pole};

		\begin{axis}[
			width=12cm,
			height=4cm,
			at={(-0.0, -10.0)},
			%xticklabel style={
			%    /pgf/number format/fixed,
			%    /pgf/number format/precision=3
			%},
			axis lines=none,
			scaled x ticks=false,
			%legend pos=north east,
			legend style={at={(0.95,0.87)}},
			cycle list name=black white,
			%domain=-0.006:0.0,
			%xmin=-0.006,
			%xmax=0.0
			]
			\addplot[only marks] table {data/nodes_top.txt};
			\addplot[only marks, mark=triangle*] table {data/nodes_fixed.txt};
			\addplot[only marks, mark=square*] table {data/nodes_internal.txt};
			\legend {$\in\mathcal{C}$, $\in\mathcal{F}$, $\in\mathcal{N}$};
		\end{axis}

		\draw[thick,->] (4.21-3.5,2.7) -- (5.21-3.5,2.7) node[anchor=south west] {$x$};
		\draw[thick,->] (4.21-3.5,2.7) -- (4.21-3.5,1.7) node[anchor=north east] {$z$};
		\draw[thick, fill=white] (4.21-3.5,2.7) circle (0.1) node[anchor=south east] {$y$};
		\draw[fill=black] (4.21-3.5, 2.7) circle (0.02);

		%\node at (-0.93, 2.99) {
		\begin{axis}[
			width=12cm,
			height=4cm,
			y dir=reverse,
			at={(-0.0, -745.0)},
			%xticklabel style={
			%    /pgf/number format/fixed,
			%    /pgf/number format/precision=3
			%},
			axis lines=none,
			scaled x ticks=false,
			%legend pos=north east,
			legend style={at={(1.0,1.0)}},
			cycle list name=black white,
			%domain=-0.006:0.0,
			%xmin=-0.006,
			%xmax=0.0
			]
			\addplot[only marks] table {data/nodes_top.txt};
			\addplot[only marks, mark=triangle*] table {data/nodes_fixed.txt};
			\addplot[only marks, mark=square*] table {data/nodes_internal.txt};
		\end{axis}
		%};
		
		%\node at (16.0, 5.0){ };
		
		\draw[thick, ->] (5.21+0.87, 3.23) -- (5.21+0.87+0.15, 3.8);
		\node at (5.21+0.87+0.4, 4.0) {$F_{i}$};
		\node at (5.21+0.87+0.4, 1.3) {$\tilde{F}_{i}$};
		\draw[thick, ->] (5.21+0.87, 2.17) -- (5.21+0.87+0.15, 1.6);
		%node[anchor=north west] {$\tilde{F}_{i,k}$};;
		\draw[dashed] (5.21+0.87, 3.23) -- (5.21+0.88, 2.17);
		\draw[thick, ->] (5.21+0.87, 2.7) -- (5.21+0.87-0.4, 2.7);
		\node at (5.21+0.87+0.54, 2.5+0.15) {$F_{\text{p},i}$};
		\draw[thick] (5.21+0.87, 2.7+0.1) -- (5.21+0.87, 2.7-0.1);
		
		% Sensor texts
		\node at (5.21, 2.7+3.25) {Sensor 1};
		\node at (5.21, 2.7-3.25) {Sensor 2};
		
	\end{tikzpicture}
	\caption{The nodes of the finite element mesh are assigned to three sets: $\mathcal{C}$ if in contact with the pole, $\mathcal{F}$ if constrained to not move, $\mathcal{N}$ otherwise.}
	\vspace{-0.5cm}
	\label{fig:fem}
\end{figure}

As mentioned above, the contact forces are postulated to be the superposition of forces $F_{i}^0$ arising from the normal indentation of the pole into the sensor, and the lateral friction forces $F_i^f$, that is,  $F_{i}=F_{i}^0 + F_{i}^\text{f}$, which implies
\begin{align}
  % {\scriptstyle\implies}
  F_{\text{p},i}=\underbrace{-\left(F_{i}^0 + \tilde{F}_{i}^0 \right)}_{=:F_{\text{p},i}^0} \underbrace{-\left(F_{i}^\text{f} + \tilde{F}_{i}^\text{f} \right)}_{=:F_{\text{p},i}^\text{f}} \label{eq:fi}
\end{align}
%$F_{i}^0$ denotes the force acting at node $i$ when no lateral friction is present. It is the result of a pure vertical indentation through contact with the pole. On the other hand, $F_{i}^\text{f}$ denotes the lateral friction that arises from the normal component of $F_{i}^0$. \textcolor{red}{The analogues..}

%Reaction force also same ...
%\begin{align}
%    F_{p,i}&=F_{p,i}^0 + F_{p,i,k}^f \ . \label{eq:fi}
%\end{align}

%\subsubsection{Contact Resolution}
\subsubsection{Forces Arising from Normal Indentation}
The forces $F_{i}^0$ are derived using the finite element theory for linearly elastic materials. Let $U_i^0$ be the deformation of a node $i$. Then, a linear relationship between the external forces and deformations is found by the finite element method as: %Use linear elastic material model to get relationship
\begin{align}
\label{eq:linear_elasticity}
    F^0=K U^0
\end{align}
where
    $F^0 := \begin{pmatrix}
    F_{1}^0,\ \dots,\ F_{N}^0  % F_{2}^0 &
    \end{pmatrix}$, $
    U^0 := \begin{pmatrix}
    U_{1}^0,\ \dots,\ U_{N}^0 %& U_{2}^0 
    \end{pmatrix}$,
and $K$ is the global stiffness matrix, obtained in this work in Abaqus/Standard.
%\textcolor{red}{.. If for each node either the deformation or force is known, a system of linear equations is obtained that can be solved for the remaining forces and deformations...} using the sets introduced above $U_i=\mathbf{0} \forall i \in \mathcal{F}$ {\color{red} explain deformations and forces already when introducing sets?}
The system of equations in \eqref{eq:linear_elasticity} can be solved by introducing the following sets, displayed in Fig.~\ref{fig:fem}:
\begin{itemize}
	\item \hl{$\mathcal{C}$: The set of all nodes that are in contact with the pole. It is the intersection of the set of nodes at the surface of the sensor (i.e., the set $\mathcal{S}$) and the set of nodes whose positions at rest collide with the pole, based on the geometric properties of the pole considered. The nodes in this set are assumed here to translate only in $z$-direction, and their deformation is obtained by finding the appropriate $z$-coordinate  that intersects with the surface of the pole (see Fig.~{\ref{fig:fem}}). }
	\item $\mathcal{F}$: The set of all nodes that are in contact with the base layer of the sensor. Since the base layer's stiffness is much larger than the stiffness of the sensor surface, the nodes of this set are assumed to be rigid. Therefore, their deformation is set to zero, i.e., $U_i^0 = 0, \forall i \in \mathcal{F}$.
	\item $\mathcal{N}$: The set of nodes that are neither in contact with the base layer nor in contact with the pole. No external forces are acting on these nodes, i.e., $F_i^0 = 0, \forall i \in \mathcal{N}$.
\end{itemize}
Therefore, for a node $i$, once a corresponding set is identified, either the force $F_i^0$ or the deformation $U_i^0$ is known. The system \eqref{eq:linear_elasticity} is then solved by using the UMFPACK library, and $F_{\text{p},i}^0$ computed as in \eqref{eq:fi}.

\hl{Note that here the current approach exploits the cylindrical geometry of the poles, rendering a mathematically simple intersection problem, which enables a highly efficient identification of the aforementioned sets. The extension to objects of various geometries may still be addressed efficiently by employing algorithms tailored to solve the intersection problem for generic polygons, e.g., based on the Weiler-Atherton clipping algorithm {\cite{weiler1977hidden}}.}
%These sets are displayed in \cref{fig:fem}.

%Friction forces ..
\subsubsection{Lateral Friction Forces}
In order to find the lateral friction forces for the nodes in contact, first the case where only the friction at a single node is unknown is considered. %{\color{red}There, the friction force is derived from the relative velocity at the subsequent timestep (semi-implicit).} 
From the solution of this case, an iterative method is utilized to solve for all friction forces in the multi-contact case.

For this, it is assumed that all the friction forces except for the one at node $i$ are known. That is $F_{j}^\text{f}$ is known for all $j\neq i$. % From \cref{eq:fi}, it follows that $F_{j,k}$ is known as well, and from \cref{eq:fpik} we also know $F_{p,j,k}$, i.e. the resultant force acting on the pole at node $j$.
%As described above, we solve for the lateral friction force $F_{i}^\text{f}$ acting on the node $i$ in a semi-implicit way. Note that it is easier to solve for the resultant friction $F_{\text{p},i}^\text{f}$. Since $F_{\text{p}}^\text{f}$ is directly related to $F_{i}^\text{f}$ through \cref{eq:equiv}.., one variable can be directly deduced from the other.  
Let
\begin{align}
    \mathbf{v}_{i,\text{rel}} &:= 
        \begin{pmatrix}
        \dot{x}_{i}\\
        \dot{y}_{i}
    \end{pmatrix} -
    \begin{pmatrix}
        \dot{x}_{\text{s}}\\ 0
    \end{pmatrix} 
    =\begin{pmatrix}
        \dot{x}_{\text{p}}\\
        \dot{y}_{\text{p}}
    \end{pmatrix}+
    \begin{pmatrix}
        -\mathbf{r}_{i,y} \\
        \mathbf{r}_{i,x}
    \end{pmatrix}
    \dot{\phi}
     -
    \begin{pmatrix}
        \dot{x}_{\text{s}}\\ 0
    \end{pmatrix} \nonumber
\end{align}
be the relative planar velocity of the point on the pole which is in contact with the node $i$ at time $k$. %Note the slight abuse of notation for the cross product, since $\dot{\phi}_{k+1}$ is a scalar and not a vector. We interpret it as the first two elements of the vector 
%\begin{align}
%    \begin{pmatrix}
%    0 \\ 0 \\ \dot{\phi}_k
%\end{pmatrix} \times \mathbf{r}_i \quad .
%\end{align}
%Then, for the next timestep, the relative velocity is given by
Then, by plugging in the equations of motion \eqref{eq:eomfirst}-\eqref{eq:eomlast}, the relative velocity at the next timestep is found to be
\begin{align}
    \mathbf{v}_{i,\text{rel}}^+&=
    \mathbf{v}_{i,\text{rel}}\!-\!\Delta t 
    \begin{pmatrix}
        \ddot{x}_\text{s} \\0
    \end{pmatrix}+\mathbf{J}_{ii} \underbrace{F_{\text{p},i,x:y}}_{\mathclap{F_{\text{p},i,x:y}^0+F_{\text{p},i,x:y}^\text{f}}}+\sum\limits_{j\neq i} \mathbf{J}_{ij} F_{\text{p},j,x:y} \label{eq:vrel}
\end{align}
where, for generic indices $a$ and $b$,
\begin{align}
    \mathbf{J}_{ab} = \Delta t\begin{bmatrix}\dfrac{1}{m_\text{p}}+\dfrac{\mathbf{r}_{a,y}\mathbf{r}_{b,y}}{I_\text{p}} & \dfrac{-\mathbf{r}_{a,y}\mathbf{r}_{b,x}}{I_\text{p}}\\% & 0\\ 
    \dfrac{-\mathbf{r}_{a,x}\mathbf{r}_{b,y}}{I_\text{p}} & \dfrac{1}{m_\text{p}}+\dfrac{\mathbf{r}_{a,x}\mathbf{r}_{b,x}}{I_{\text{p}}}  \end{bmatrix}.
\end{align}
In this work, Coulomb friction is assumed, and two cases are identified, where $\mu$ indicates both the static and kinetic friction coefficients.
%Using the Coulomb friction model, two cases
%bound by .. normal component of the force, i.e. $|F_{i,z}| + |\tilde{F}_{i,z}|$
%\textcolor{red}{Find lateral friction force.. how? .. Coulomb friction model states that friction cone.., differentiate between static friction and kinetic friction. }
First, for the static friction case, consider the force $F_{\text{p},i}^{\text{f},\text{static}}$ that takes on exactly the value to prevent motion at node $i$. This force can be found by setting \mbox{$\mathbf{v}_{i,\text{rel}}^+=\mathbf{0}$} in \eqref{eq:vrel} and solving for $F_{\text{p},i,x:y}^f$. If this force satisfies the friction cone constraint, i.e.
\begin{align}
\label{eq:static_friction}
    %\text{if } 
    \mynorm{F_{\text{p},i}^{\text{f},\text{static}}}\leq 2\mu \left|F_{i,z}^0\right|, %\text{then } F_{\text{p},i}^\text{f}=F_{\text{p},i}^{\text{f},\text{static}}
\end{align}
then $F_{\text{p},i,x:y}^\text{f}=F_{\text{p},i,x:y}^{\text{f},\text{static}}$.
The $z$-component is set to zero as it would eventually cancel out when considering both fingers.

%First, for the static friction case, that is, when 
%\begin{align}
%\label{eq:static_friction}
%    %\text{if } 
%    \mynorm{F_{\text{p},i}^{\text{f},\text{static}}}\leq 2\mu \left|F_{i,z}^0\right|, %\text{then } F_{\text{p},i}^\text{f}=F_{\text{p},i}^{\text{f},\text{static}}
%\end{align}
%the force takes on exactly the value to prevent motion, i.e. $\mathbf{v}_{i,\text{rel}}^+=\mathbf{0}$, then 
%\begin{align}
    %F_{\text{p},i,x:y}^{\text{f},\text{static}}=-\mathbf{J}_i^{-1}\Big(&\mathbf{v}_{i,\text{rel}}-\Delta t    \begin{pmatrix}
    %    \ddot{x}_\text{s} \\0
    %\end{pmatrix} +\mathbf{J}_i F_{\text{p},i,x:y}^0
    %+\sum_{j\neq i} \mathbf{J}_j F_{\text{p},j,x:y} \Big)  \nonumber%TODO \nonumber
%\end{align}
%the $x$-$y$ components of $F_{\text{p},i}^\text{f}$ can be computed by solving \eqref{eq:vrel} for $F_{\text{p},i,x:y}^\text{f}$. The $z$-component is set to zero as it would eventually cancel out when considering both fingers.
If \eqref{eq:static_friction} is not satisfied, friction is not sufficient to prevent the motion at node $i$. In this case, kinetic friction is present, where the force is opposite to the direction of the velocity and is proportional to the normal component of the force. %Again, using $v_{i,\text{rel}(k)}$ yields inconsistent.. that's why semi-implicit..
Since the velocity ${\mathbf{v}}_{i,\text{rel}}^+$ and the friction force $F_{p,i}^\text{f}$ are coupled, an approximation of the subsequent velocity is employed as
\begin{align}
    \hat{\mathbf{v}}_{i,\text{rel}}^+&:=
    \mathbf{v}_{i,\text{rel}}-\Delta t 
    \begin{pmatrix}
        \ddot{x}_\text{s} \\0
    \end{pmatrix}+\mathbf{J}_i {F^0_{\text{p},i,x:y}}+{\sum}_{j\neq i} \mathbf{J}_j F_{\text{p},j,x:y} \nonumber
\end{align}
which is the relative velocity at the subsequent step when the effects of friction at node $i$ are ignored. If the number of nodes is sufficiently large, this approximation is close to the true value, since the effect of the force at the single node $i$ is small compared to the combined effect of the remaining forces at nodes $j\neq i$. Using this approximation, the kinetic friction is set to
\begin{align}
    %\text{if}\, \mynorm{F_{\text{p},i}^{\text{f},\text{static}}}> 2\mu|F_{i,z}^0|,\, \text{then}\,
    F_{\text{p},i,x:y}^\text{f}=- %\frac{\tilde{\mathbf{v}}_{i,\text{rel}}^+}{a}
    2\mu|F_{i,z}^0|{\hat{\mathbf{v}}_{i,\text{rel}}^+}/{\mynorm{\hat{\mathbf{v}}_{i,\text{rel}}^+}}  %\label{eq:kineticf} 
\end{align}

%Finally, iteratively solve ..iterate over each contact and solve as single contact problem .. put reference here..
Given this solution to the single contact problem, the multi-node contact case is solved by repeatedly iterating over all nodes in contact until convergence, and updating the friction force at node $i$ using the above solution, given the values at nodes $j\neq i$ of the current iteration.
Then, $F_{\text{p},i}$ is obtained as in \eqref{eq:fi} from $F_{\text{p},i}^0$ and $F_{\text{p},i}^\text{f}$, and finally $F_\text{p}$ and $T_\text{p}$ can be computed as in \eqref{eq:t_tot}.

\subsection{Learning Tactile Control Policies}
\label{sec:control}

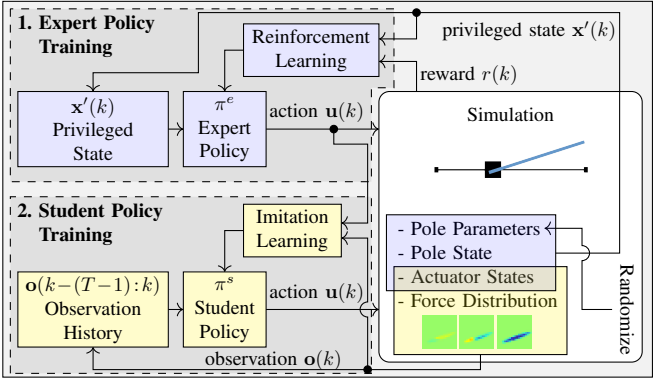
\begin{figure}
	\centering
		\begin{tikzpicture}[every node/.style={scale=0.75}]
		\setlength{\fboxsep}{0pt}
			
			% Grey boxes
			\draw[fill=black!5!white] (0, 0.2) rectangle (\columnwidth, -4.8);
			%\draw[fill=red!10!white] (0, -5) rectangle (\textwidth, -9.5);
			
			% Student Box
			\draw[dashed, fill=black!12!white] (0.1, -4.75) rectangle (4.9, -2.4);
			\node[anchor=north west, text width= 4cm] at (0.1, -2.4) {\textbf{2. Student Policy\\\phantom{2. }Training}};
			
			% Expert Box
			\draw[dashed, fill=black!12!white] (0.1, 0.1) -- (0.1, -2.2) -- (4.9, -2.2) -- (4.9, -0.95) -- (5.2, -0.95) -- (5.2, 0.1) -- cycle;
			\node[anchor=north west, text width= 4cm] at (0.1, 0.1) {\textbf{1. Expert Policy\\\phantom{1. }Training}};
			
			% Simulation
			\draw[rounded corners, fill=white] (5, -1.0) rectangle (8.5, -4.6);
			\node at (6.75, -1.3) {Simulation};
			\node[inner sep=0] at (6.75, -2.05) {{\includegraphics[width=0.35\columnwidth]{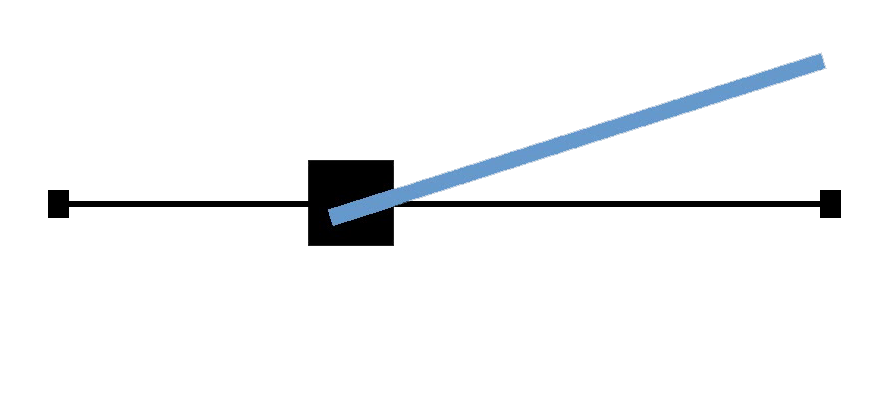}}};
			
			% Reinforcement Learning
			\path[fill=white] (3.2, -0.1) rectangle (5.0, -0.8);
			\draw[fill=blue, fill opacity=0.1, text opacity=1.0] (3.2, -0.1) rectangle (5.0, -0.8) node[midway, text width=2.2cm, text centered] {Reinforcement Learning};
			\node at (6.2, -0.775) {reward $r(k)$};
			\draw[->] (5.5, -1.0) -- (5.5, -0.6) -- (5.0, -0.6);
			\draw[->] (3.2, -0.45) -- (2.95, -0.45) -- (2.95, -1.0);
			\draw[fill=black] (5.5, 0.05) circle (0.05);
			\draw[->] (5.5, 0.05) -- (5.5, -0.3) -- (5.0, -0.3);
			
			% States
			\node[text width=4cm] at (6.75, -3.3) {- Pole Parameters\\- Pole State\\- Actuator States\\- Force Distribution};
			\node[rotate=-90] at (8.3, -3.9) {Randomize};
			\draw[->] (8.1, -3.9) -- (7.7, -3.9) -- (7.7, -3.25) arc(90:270:-0.1) -- (7.7, -2.82) -- (7.2, -2.82);
			
			% Force distribution
			\node[inner sep=0] at (5.8, -4.2) {\includegraphics[width=0.6cm]{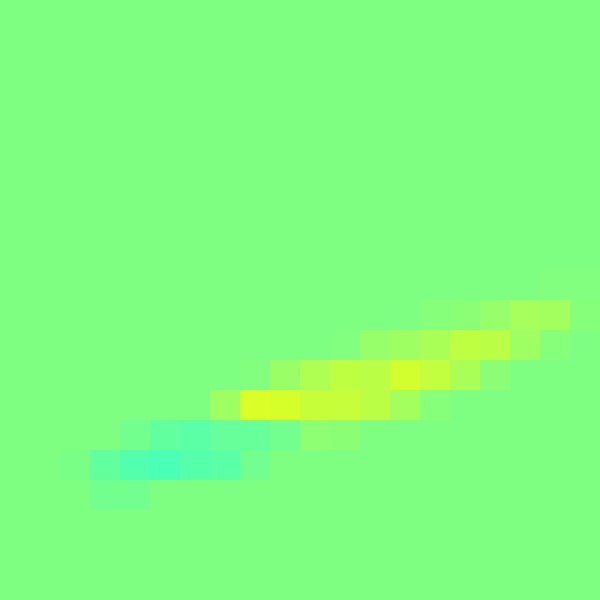}};
			\node[inner sep=0] at (6.3, -4.2) {\includegraphics[width=0.6cm]{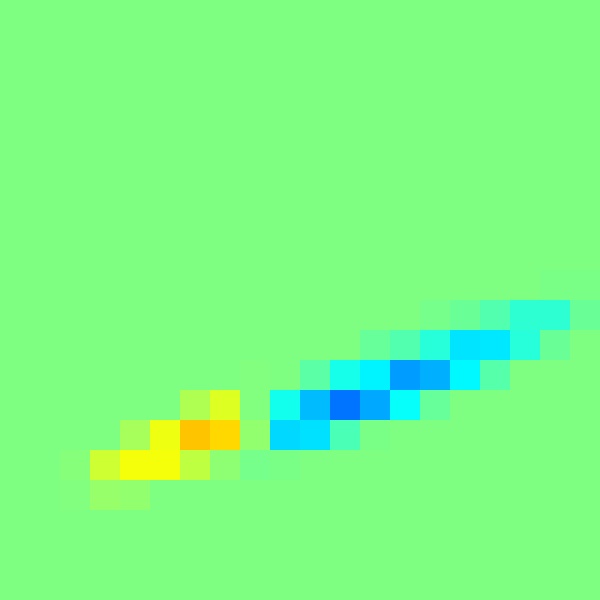}};
			\node[inner sep=0] at (6.8, -4.2) {\includegraphics[width=0.6cm]{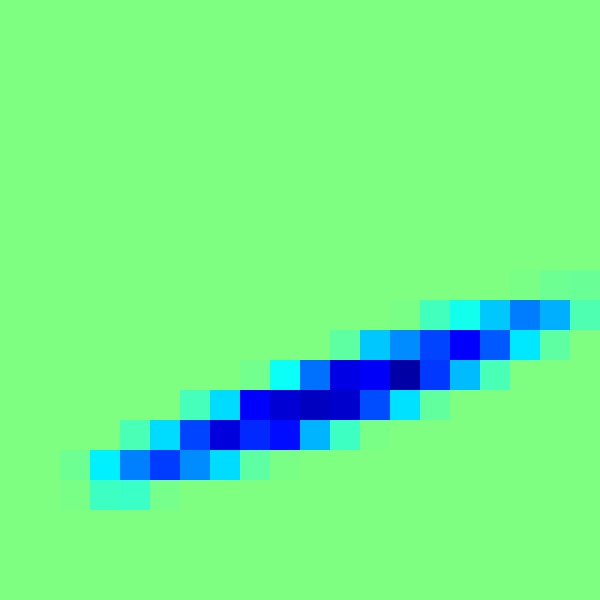}};
			
			% Student Policy
			\draw[fill=yellow, fill opacity=0.2] (5.2, -4.5) rectangle (7.5, -3.33);
			\draw[->] (6.35, -4.5) -- (6.35, -4.7) -- (1.2, -4.7) -- (1.2, -4.4);
			\path[fill=white] (2.4, -3.4) rectangle (3.5, -4.4);
			\draw[fill=yellow, fill opacity=0.2, text opacity=1.0] (2.4, -3.4) rectangle (3.5, -4.4) node[midway, text width=2cm, text centered] {$\pi^s$\\Student Policy};
			\draw[->] (3.5, -3.9) -- (5, -3.9) node[xshift=-1.1cm, above] {action $\mathbf{u}(k)$};
			\draw[fill=black] (4.85, -4.7) circle (0.05);
			\draw[->] (4.85, -4.7) -- (4.85, -4.0) arc(-90:-270:0.1) -- (4.85, -2.95) -- (4.5, -2.95);
			
			% Observation History
			\path[fill=white] (0.2, -3.4) rectangle (2.2, -4.4);
			\draw[fill=yellow, fill opacity=0.2, text opacity=1.0] (0.2, -3.4) rectangle (2.2, -4.4) node[midway, text width=2.5cm, text centered] {$\mathbf{o}(k\!-\!(T\!-\!1)\!:\!k)$\\Observation History};
			\node at (3.6, -4.575) {observation $\mathbf{o}(k)$};
			\draw[->] (2.2, -3.9) -- (2.4, -3.9);
			
			% Privileged State
			\path[fill=white] (0.2, -1.0) rectangle (2.2, -2.0);
			\draw[fill=blue, fill opacity=0.1, text opacity=1.0] (0.2, -1.0) rectangle (2.2, -2.0) node[midway, text width=1.9cm, text centered] {$\mathbf{x}'(k)$\\Privileged State};
			\node at (7.0, -0.2) {privileged state $\mathbf{x}'(k)$};
			\draw[->] (2.2, -1.5) -- (2.4, -1.5);
			
			% Expert Policy
			\draw[fill=blue, fill opacity=0.1] (5.1, -3.65) rectangle (7.35, -2.64);
			\draw[->] (7.35, -3.15) -- (8.2, -3.15) -- (8.2, 0.05) -- (2.6, 0.05) -- (2.6, -0.75) -- (1.2, -0.75) -- (1.2, -1.0);
			\path[fill=white] (2.4, -1.0) rectangle (3.5, -2.0);
			\draw[fill=blue, fill opacity=0.1, text opacity=1.0] (2.4, -1.0) rectangle (3.5, -2.0) node[midway, text width=1.7cm, text centered] {$\pi^e$\\Expert Policy};
			\draw[->] (3.5, -1.5) -- (5, -1.5) node[xshift=-1.1cm, above] {action $\mathbf{u}(k)$};
			\draw[fill=black] (4.4, -1.5) circle (0.05);
			\draw[->] (4.4, -1.5) -- (4.4, -1.7) -- (4.85, -1.7) -- (4.85, -2.75) -- (4.5, -2.75);

			% Imitation Learning
			\path[fill=white] (3.2, -2.5) rectangle (4.5, -3.2);
			\draw[fill=yellow, fill opacity=0.2, text opacity=1.0] (3.2, -2.5) rectangle (4.5, -3.2) node[midway, text width=2cm, text centered] {Imitation Learning};
			\draw[->] (3.2, -2.85) -- (2.95, -2.85) -- (2.95, -3.4);

		\end{tikzpicture}
	\caption{\hl{The figure depicts the privileged learning approach. The expert and student policy training take place in two separate steps, both performed entirely in simulation. Further, both policies are parametrized using two-layer fully connected neural networks. For the reinforcement learning of the expert policy, the SAC {\cite{sac}} method is employed to find the optimal policy according to ({\ref{eq:policy}}). The student policy is then deployed to the real-world system without further adaptation.}}
	\vspace{-0.5cm}
	\label{fig:imitation}
\end{figure}

Given a simulation of a robotic system, deep reinforcement learning algorithms have been successfully applied to learn sophisticated behaviours \cite{kober2013reinforcement}.
These algorithms typically depend on the Markov property of the system, i.e. they assume that the state and physical parameters that fully describe the system at a given time are available to the policy. However, for the experiments presented in \cref{sec:results}, the physical parameters of the pole, e.g. the length, are unknown to the policy and the Markov property no longer holds. In reinforcement learning, such problems are typically dealt with by using the history of observations and parametrizing the policy using recurrent neural networks. However, such approaches can be challenging to train.% and typically require many more interactions with the environment when compared to the policies that learn from the complete state.

The approach employed here exploits the fact that in simulation the state and physical parameters are known. In a first stage, an \emph{expert} policy $\pi^\text{e}$ is learned that has access to the state as well as the simulation parameters. In a second stage, a \emph{student} policy $\pi^\text{s}$ that only has access to the observations that are available on the real system is learned by imitating the behaviour of the expert policy (see \cref{fig:imitation}). This idea is also referred to as privileged learning \cite{chen2020learning, lee2020learning}.

\subsubsection{State-Feedback Expert Policy}

In order to achieve the swing-up with a feedback policy that adapts to different poles, the expert policy is conditioned on the state $\mathbf{x}(k)$ (defined in \cref{sec:sim}), as well as the pole's physical parameters which may vary. This yields the augmented state
\begin{align}
	\mathbf{x}'(k) &:= \begin{pmatrix}
		\mathbf{x}(k),\ 
		r_\text{p},\ 
		m_\text{p},\ 
		I_\text{p},\ 
		l_{\text{p},\text{u}},\ 
		l_{\text{p},\text{l}},\ 
		\mu
	\end{pmatrix} \quad .
\end{align}
As a result, the policy may choose different control actions based on the features of the pole.

The goal is then to find a policy, $
	\pi \colon {\mathbf{x}'}(k) \to \mathbf{u}(k)
$,
that is optimal in the sense of maximizing the expected sum of future discounted rewards, i.e.
\begin{align}
	\pi^\text{e} &= \max_{\pi} \mathbb{E}\left( \sum_k{\gamma^k r({\mathbf{x}'}(k), \pi({\mathbf{x}'}(k))) } \right) \quad , \label{eq:policy}
\end{align}
where $\gamma$ is the discount factor. The reward function $r$ is shaped to encourage low slippage and pole orientations that are close to $\SI{180}{\degree}$.
%\begin{align}
%    r(\hat{\mathbf{x}}_k, \mathbf{u}_k) &=
%    \exp\left(- \left|\frac{\phi_k-\pi}{0.7}\right| \right)
%    \left(\exp\left(- \left(\frac{x_{p,\text{pivot},k}-x_{s,k}}{0.04}\right)^{2} \right)
%    \exp\left(- \left(\frac{y_{p,\text{pivot},k}}{0.04}\right)^{2} \right)
%    + \exp \left(-\left( \frac{\dot{x}_s}{0.3}\right)^2 \right)\right)\quad . \label{eq:reward}
%\end{align}
%\begin{align}
%	r(\hat{\mathbf{x}}_k, \mathbf{u}_k) &=
%	\exp\left(- \left|\frac{\phi_k-\pi}{0.7}\right| \right)
%	\exp\left(- \left(\frac{x_{p,\text{pivot},k}-x_{s,k}}{0.04}\right)^{2} \right)
%	\exp\left(- \left(\frac{y_{p,\text{pivot},k}}{0.04}\right)^{2} \right)\quad . \label{eq:reward}
%\end{align}
%({\color{red} need to compactly define reward and terminal states})
% \begin{table}[]
% 	\centering
% 	\renewcommand{\arraystretch}{1.3}
% 	\begin{tabular}{llll} \toprule
% 		Parameter & Unit & Description & Distribution \\ \midrule
% 		$l_{\text{p},\text{u}}$ & $\SI{}{\metre}$& Length above CoM & $\mathcal{U}(0.100, 0.250)$\\
% 		$l_{\text{p},\text{l}}$ & $\SI{}{\metre}$& Length below CoM & $\mathcal{U}(0.050, 0.100)$\\
% 		$m_\text{p}$ & $\SI{}{\kilo\gram}$ & Mass & $\mathcal{U}(0.010, 0.035)$\\
% 		$\mu$ & $-$ & Coeff. of friction & $\mathcal{U}(0.450, 0.600)$\\ 
% 		$I_\text{p}$ & \SI{}{\kilo\gram\metre\squared}& Moment of inertia & 
% 		\vtop{\hbox{\strut $\mathcal{U}(1/12\ l_\text{p}^2\ m_\text{p},$
% 			}\hbox{\strut $\phantom{\mathcal{U}(}1/4\ l_\text{p}^2\ m_\text{p})$}}
% 		\\\bottomrule
% 	\end{tabular}
% 	\caption{Simulation parameters}
% 	\vspace{-1cm}
% 	\label{tab:params}
% \end{table}
The policy is learned using deep reinforcement learning, namely the SAC \cite{sac} algorithm with the stable-baselines3 implementation \cite{stable-baselines3}. The discount factor is set to $\gamma=0.995$ while the remaining hyperparameters, as well as the policy network architecture, correspond to the default ones proposed in \cite{sac}.
%Note that the output layer contains a $\tanh$ activation function that scales the control actions to the appropriate range. The range for the cart acceleration $\ddot{x}_{\text{s}}$ is fixed to $[\SI{-40}{\metre\per\second\squared}, \SI{40}{\metre\per\second\squared}]$ and the commanded gripping position $z_{\text{s}}$ is scaled to $[-r_{\text{p}}+\SI{0.0003}{\metre}, -r_{\text{p}} + \SI{0.002}{\metre}]$. The gripping distance is constrained to this range such that there is always at least a $\SI{0.3}{\milli\metre}$ indentation in the sensor which ensures that the material deformation on the real-world sensor is sufficiently high in order to infer a sensible estimate of the force distribution in the presence of image noise. Moreover the maximum indentation is limited to $\SI{2.0}{\milli\metre}$ so as to limit the normal force acting on the pole to a reasonable value.
\hl{During training, the pole parameters are randomly sampled at each new episode such that the policy learns the correct behaviour for different poles. }%The probability distribution of the parameters is listed in \cref{tab:params}, where $\mathcal{U}(a,b)$ denotes the continuous uniform distribution on the interval $(a,b)$. 
This \emph{dynamics randomization} \cite{peng2018sim} also greatly aids in the successful transfer from simulation to reality.% ({\color{red} explain why?})
 %Moreover, in order to further increase the robustness of the policy, the initial position and orientation of the pole, and the initial position of the cart, are randomized as well.%As a result the policy learns the swing-up maneuvers for different initial states.

\subsubsection{Tactile Student Policy}
The expert policy is conditioned on privileged knowledge, only available in simulation, and can thus not be deployed on the real system, where the pole's pose and physical attributes can only indirectly be observed through the available force distribution measurements. As a result, the student policy must be able to reason over time and implicitly recover the missing state information. First, in order to condense the sensory information into a compressed representation, an estimate of the pole's orientation $\hat{\phi}(k)$ is obtained by computing the force magnitude at each bin, thresholding the magnitudes to obtain a binary image, and finally applying a Hough line transform \cite{hough}. In addition, the total sensed normal force $F_{z}^{\text{tot}}(k)$ is extracted by summing the $z$-distribution at all bins. This is motivated by the fact that the normal force yields direct information about the friction and slippage, while the angle of the pole is the main quantity to be controlled. Then, a student policy conditioned on a history of condensed representations of the observations is learned by imitating the behaviour of the expert policy.

A condensed observation at time $k$ is given by
\begin{align}
	{\mathbf{o}}(k)&=\begin{pmatrix}
		x_{\text{s}}(k),\
		z_{\text{s}}(k),\
		\hat{\phi}(k),\
		F_{z}^{\text{tot}}(k)
	\end{pmatrix} \ .
\end{align}
Note that $x_{\text{s}}(k)$ and $z_{\text{s}}(k)$ are known for the real system, and the velocity of the cart $\dot{x}_{\text{s}}(k)$ is not included, since it can implicitly be derived from the history of $x_{\text{s}}(k)$ observations.
%Further, since lines in a two-dimensional image {\color{red}can be rotated onto themselves every \SI{180}{\degree}}, ...filtered.

%\begin{align}
%	\mathbf{o}(k) &= \begin{pmatrix}
%		x_{s}(k)&
%		\dot{x}_{s}(k)&
%		z_{s}(k) &
%		F_{1}(k) &
%		%F_{2}(k) &
%		\dots &
%		F_{n}(k)
%	\end{pmatrix} \ ,
%\end{align}
%where $F_i$ denotes the sensed three-dimensional force of bin $i$. %Additionally, a single observation at a given timestep does not suffice to .... 
 %Note also that for a $20\times 20$ discretized three-dimensional force distribution, the observation at time $k$ is over $1'200$-dimensional. It stands to reason that a dimensionality reduction is required for the learning to be effective. Then, a student policy conditioned on a history of condensed representations of the observations is learned by imitating the behaviour of the expert policy.

%\paragraph{Observation Dimensionality Reduction}

%From the force distribution, an estimate of the pole's orientation $\tilde{\phi}_k$ and 
%Note also that the velocity of the cart $\dot{x}_s(k)$ is excluded since it can implicitly be derived from the history of positions.

%. ..This procedure is illustrated in \cref{fig:imitation}.

%\paragraph{Imitating the Expert Policy}
The student policy $\pi^\text{s}$ is then parametrized by a neural network that maps the history of the last $T$ condensed observations ${\mathbf{o}}(k\!-\!(T\!-\!1)\!:\!k)$ to the control action $\mathbf{u}(k)$, where \mbox{$T=12$} is the fixed history length. \mbox{The same stochastic} network as proposed in \cite{sac} is used, which outputs a squashed Gaussian distribution over the control actions. This stochasticity accomplishes a desirable smoothing of the policy. The imitation of the expert policy is then posed as a supervised learning task that minimizes the negative log-probability
\begin{align}
	\mathcal{L}:=-\log \text{Pr}\left(\pi^\text{s}\left({\mathbf{o}}(k\!-\!(T\!-\!1)\!:\!k)\right)=\pi^\text{e}\left(\mathbf{x}'(k)\right)\right) \ . \nonumber
\end{align}
In this work, the DAGGER \cite{dagger} method is employed, where the dataset is continuously aggregated with the incoming data from the training rollouts of the student policy. Labels are obtained by querying the expert policy for the visited states. At each training iteration, the student policy is updated by performing an optimization step with batches sampled from the aggregated dataset.
%The intuition behind this algorithm is that over the iterations, we are building up the set of inputs that the learned policy is likely to encounter during its execution based on previous experience

\section{RESULTS}
\label{sec:results}

\begin{figure*}
	\centering
	\begin{tikzpicture}
		%\draw (0, 0) -- (0,1);
		\node[inner sep=0pt] at (0,0)
		{\includegraphics[width=0.8\textwidth]{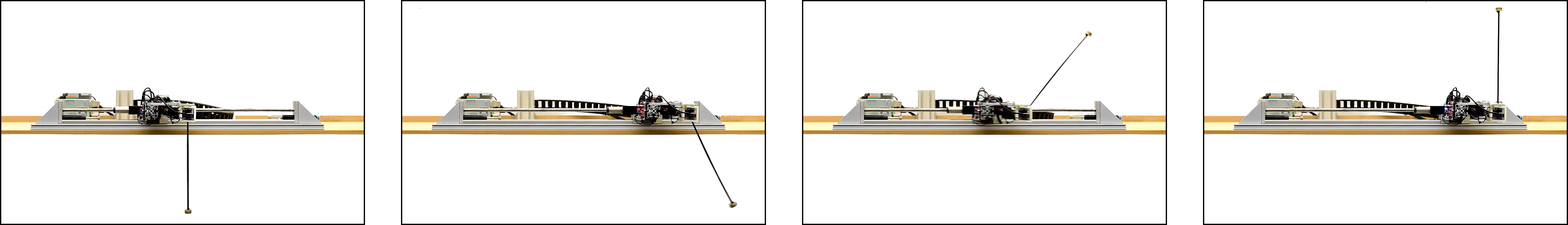}};
		%		\draw[->] (-7.5, 4.75) -- (7.5, 4.75) node[midway, above] {$t$};
		
		%		\node at (0, 3.5) {$r_p=\SI{5.0}{\milli\metre},\ l_{p,u}=\SI{0.23}{\metre},\ m_p=\SI{36}{\gram},\ I_p=\SI{3.9e-4}{\kilo\gram\metre\squared}$};
		%		
		%		\node at (0, -0.2) {$r_p=\SI{2.5}{\milli\metre},\ l_{p,u}=\SI{0.19}{\metre},\ m_p=\SI{16}{\gram},\ I_p=\SI{7.9e-5}{\kilo\gram\metre\squared}$};
		
		%		\node at (0, -4.5) {$r_p=\SI{3.9}{\milli\metre},\ l_{p,u}=\SI{0.18}{\metre},\ m_p=\SI{26}{\gram},\ I_p=\SI{1.1e-4}{\kilo\gram\metre\squared}$};
		
	\end{tikzpicture}
	\caption{This figure shows a trajectory which results from employing the learned feedback control policy on the robotic system. As can be seen, the pole is dynamically swung up to an upright position.}
	\vspace{-4mm}
	\label{fig:closedloop}
\end{figure*}

The validity of the methods presented is verified on the physical system, where the learned feedback policy is deployed to swing up different poles.
%In a first step, the accuracy of the simulation is evaluated by running the exact same control actions in both simulation and reality, and comparing the resulting trajectories and force distributions. Next, the learned feedback policy is deployed to the robotic system to swing up different poles.

%\subsection{Experimental Setup}
Feedback is crucial for this task for three reasons: i) the control actions to perform a successful swing up depend on the physical parameters of the specific pole, which are assumed to be unknown to the policy in this work, ii) these control actions depend on the initial position and orientation of the pole, which is likely to differ across trials on the real system, iii) even when the physical parameters and starting pose of the pole are well known, and a trajectory is generated in simulation for such a configuration, in the authors' experience this led to swing-ups with an offset in the final angle due to slight model mismatches. Feedback is thus needed to precisely control the final angle.

Throughout the following experiments, the initial grasping of the pole is achieved by a human holding the pole between the two tactile sensors. The gripper then slowly closes its fingers until the total force applied on the sensor by the pole reaches a user-defined threshold.

The student feedback control policy is evaluated on the real-world robotic system on four different poles with masses ranging from $\SI{20}{\gram}$ to $\SI{38}{\gram}$, lengths from $\SI{20}{\centi\metre}$ to $\SI{35}{\centi\metre}$, and radii from $\SI{2.5}{\milli\metre}$ to $\SI{5}{\milli\metre}$.
%\cref{fig:closedloop}, where the first pole is more than double the weight of the second pole ({\color{red}TBD}).
%Note that these parameters are only identified for evaluation purposes, and the student control policy does not have any knowledge about them. %The figure also contains keyframes of the trajectories that result from the policy. As can be seen,
\hl{%
For each pole, the control policy is run ten times and the error from \mbox{\ang{180}} in the final estimated angle $\hat{\phi}$ is recorded. Experiments show that all four poles are successfully swung up to an upright position, and a mean absolute error of \mbox{\ang{4.3}} is achieved. A detailed analysis of the experimental results is provided in an experimental report \mbox{\cite{experimentalreport}}.%
}
These results demonstrate how a single policy is able to adapt the robot's motion to perform swing-up maneuvers for a wide range of different poles without any prior knowledge of the pole's physical features, based on the feedback provided by the tactile sensor. The resulting behaviour of the policy for one of the listed poles is depicted in \cref{fig:closedloop}. The supplementary video\footnote{\url{https://youtu.be/In4jkaHzJLc}} contains the trajectories for the remaining poles. It is vital to note that the pole shown in \cref{fig:closedloop} is not contained in the distribution of poles that is used while learning either the teacher or the student policy.% In other words, no pole that shares the same characteristics as this one was used to train the policy.

Moreover, the policy is transferred directly from the simulation to the real system with no adaptation needed. This further asserts the robustness of the policy as it is able to adapt to the real system that exhibits dynamics that are not modeled in the simulation (such as dynamic effects of the sensor material, unmodeled dynamics of the actuators, and delays of the actuator commands).
\vspace{-1mm}
\section{CONCLUSION}
\vspace{-1mm}
\label{sec:conc}

\begin{figure}
	\centering
	\begin{tikzpicture}[scale=1.0]
		\setlength{\fboxsep}{0pt}
		\node[inner sep=0] at (0,0) {\fbox{\includegraphics[width=0.25\columnwidth]{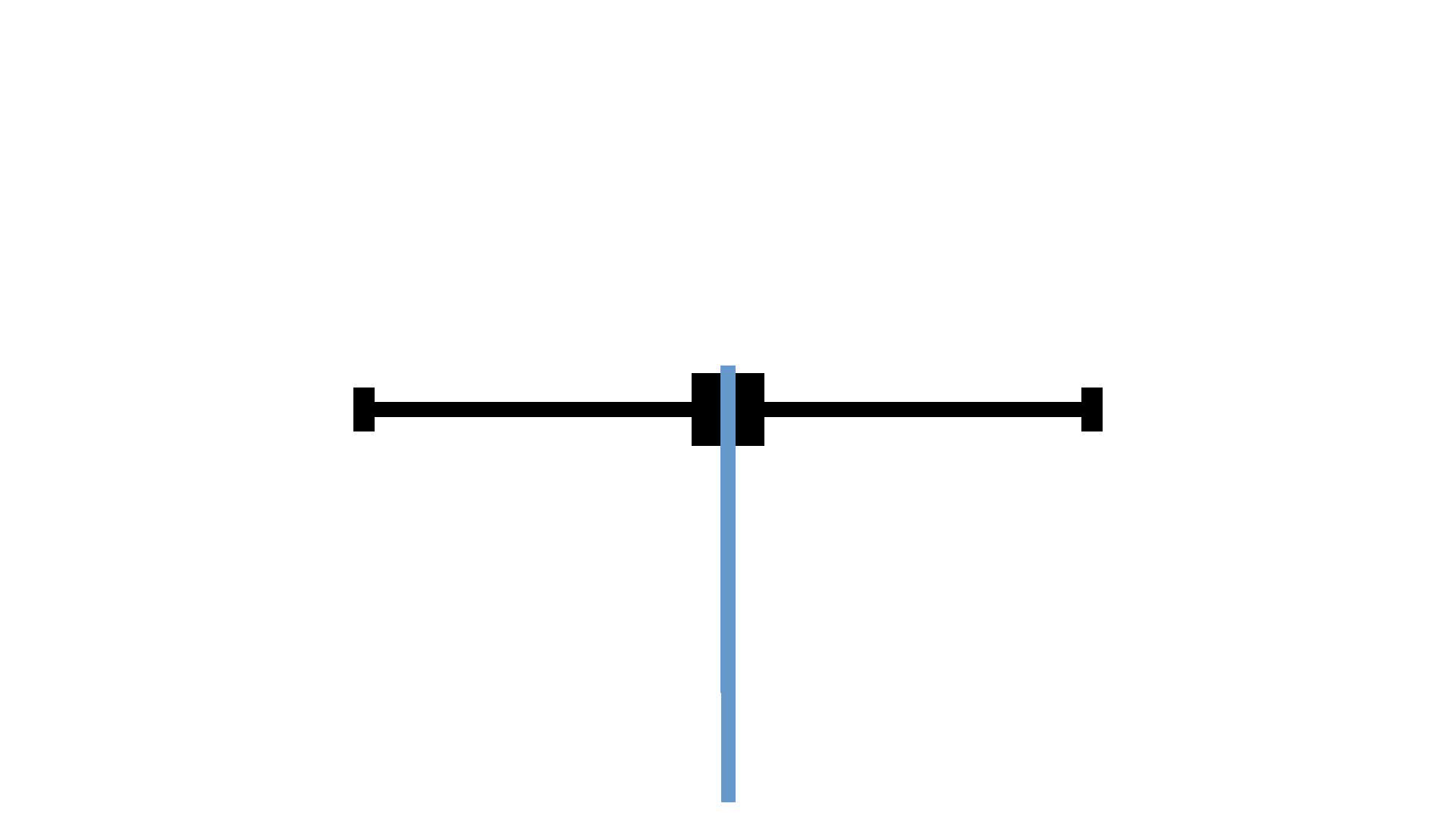}}};
		\node[inner sep=0] at (2.6,0) {\fbox{\includegraphics[width=0.25\columnwidth]{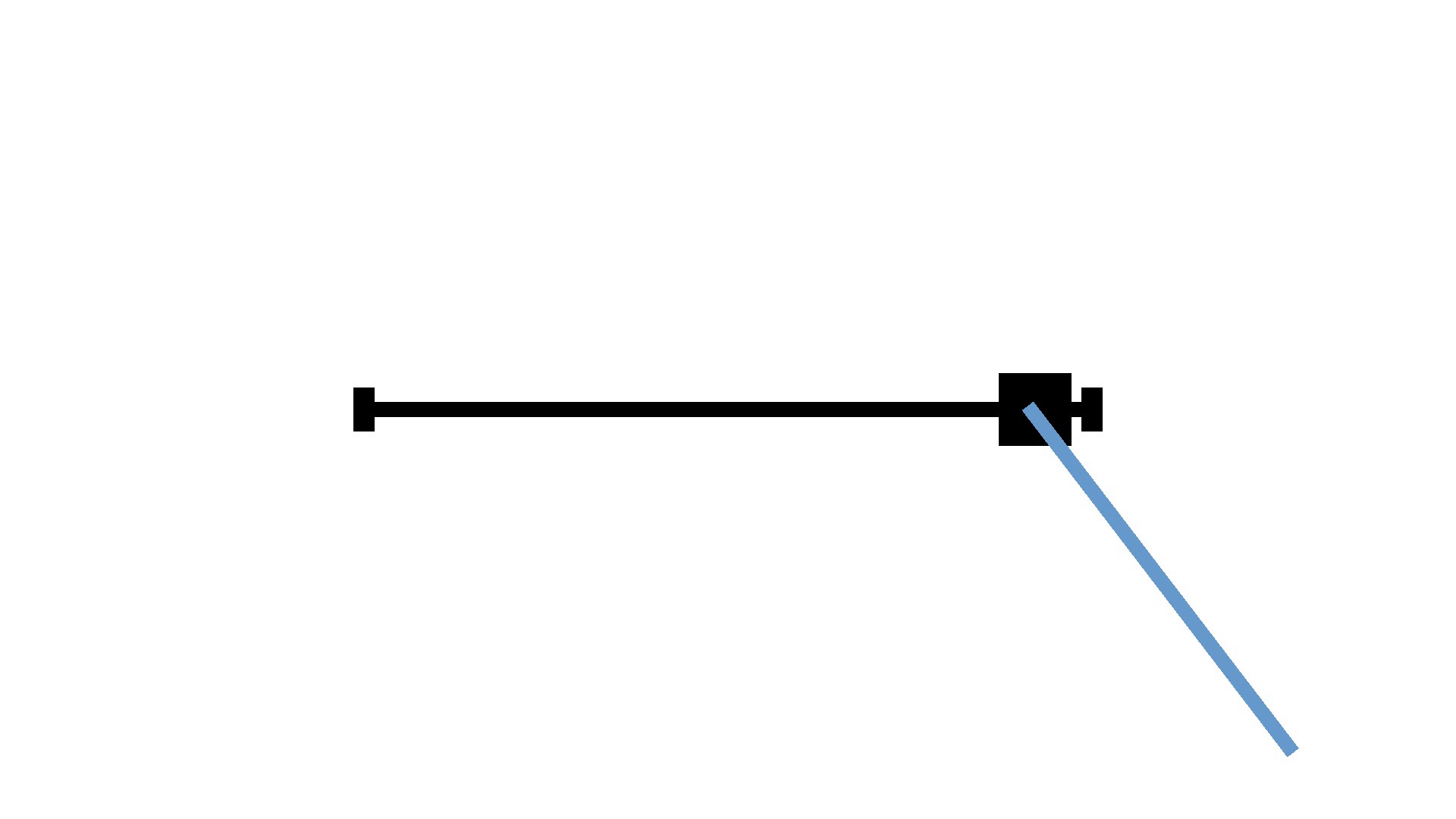}}};
		\node[inner sep=0] at (5.2, 0) {\fbox{\includegraphics[width=0.25\columnwidth]{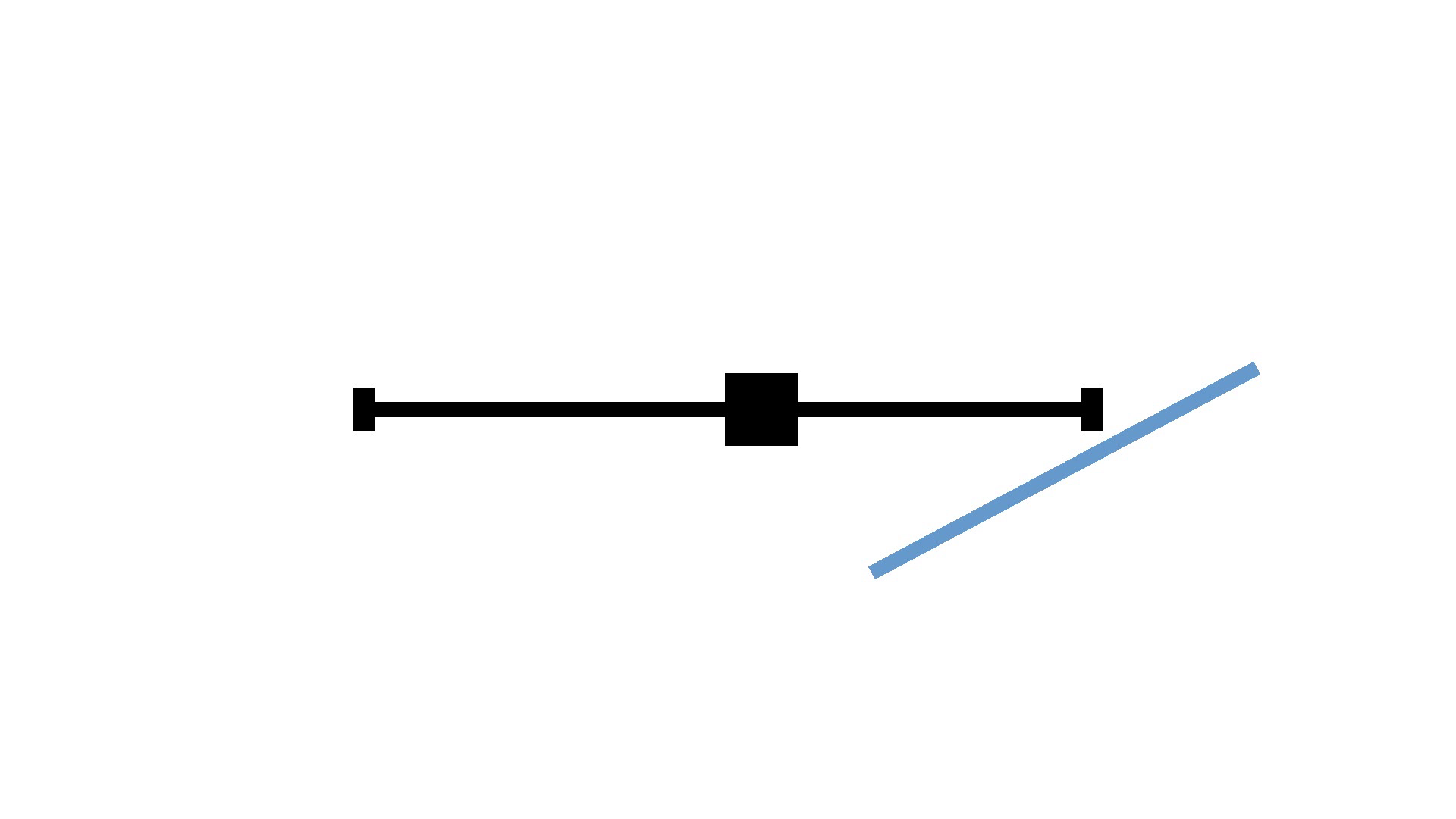}}};
		\node[inner sep=0] at (0,-1.4) {\fbox{\includegraphics[width=0.25\columnwidth]{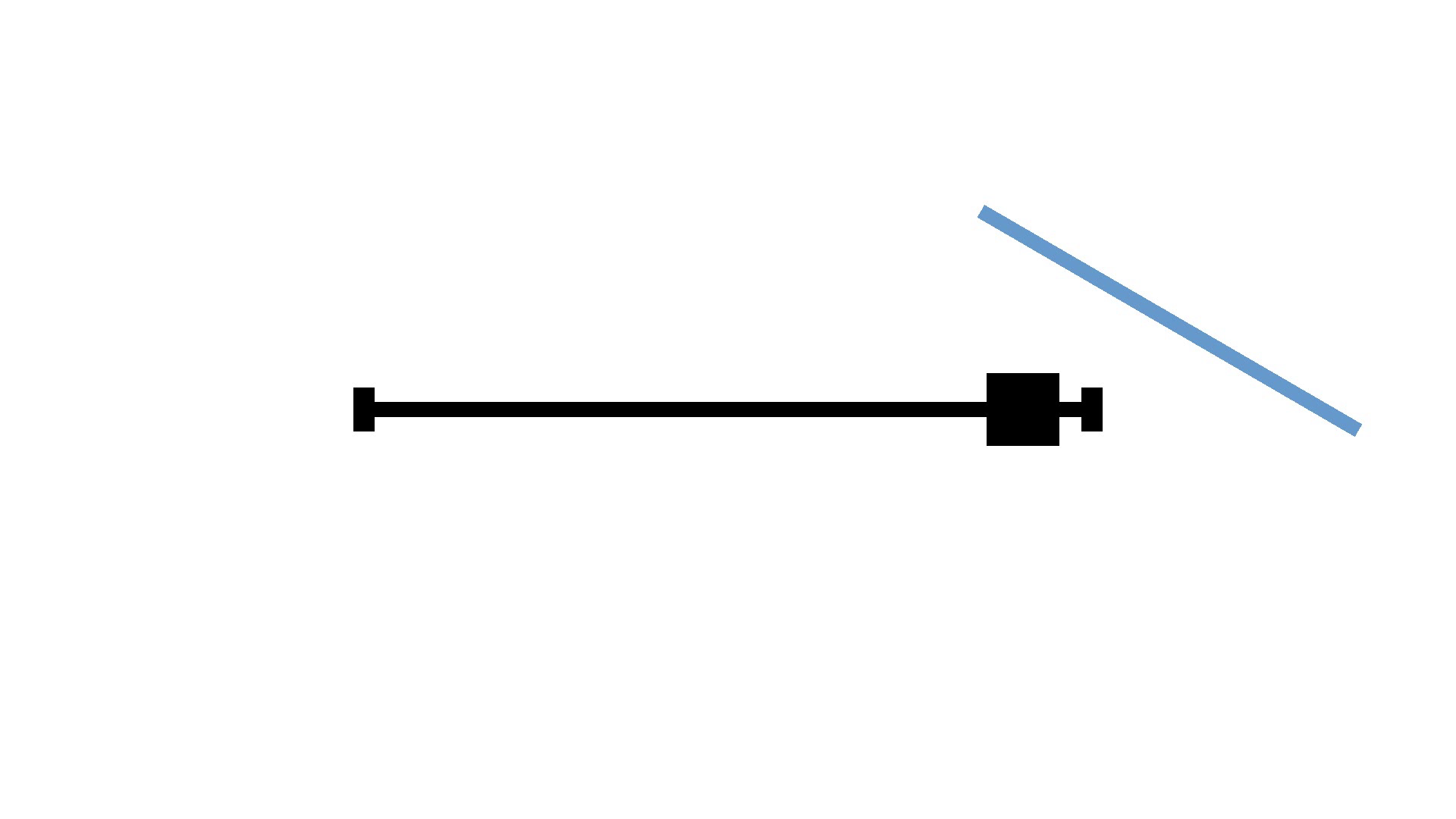}}};
		\node[inner sep=0] at (2.6,-1.4) {\fbox{\includegraphics[width=0.25\columnwidth]{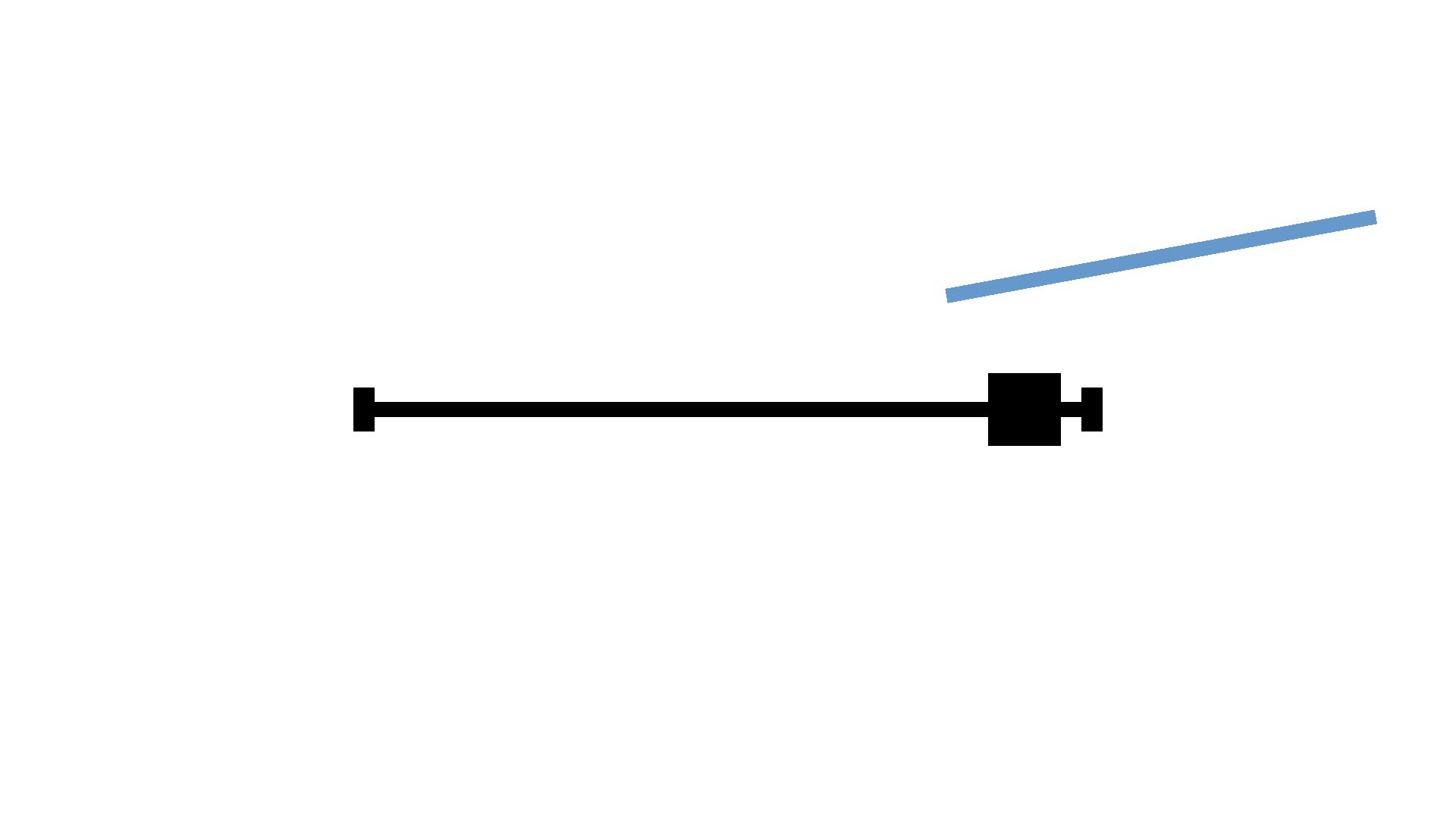}}};
		\node[inner sep=0] at (5.2,-1.4) {\fbox{\includegraphics[width=0.25\columnwidth]{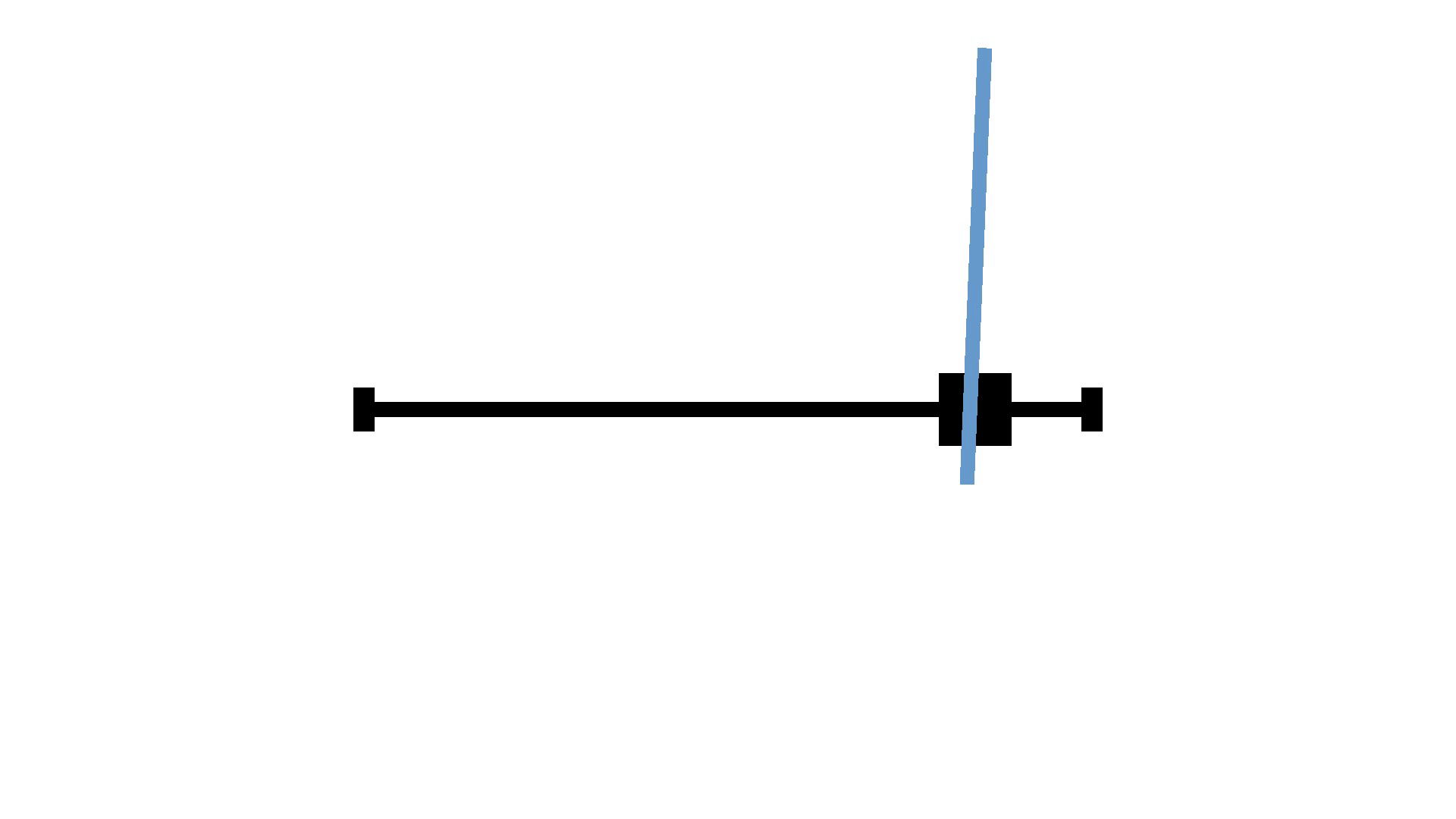}}};
		
		%\draw[->] (-1, 1.3) -- (12.5, 1.3) node[above, midway] {$t$};
		
	\end{tikzpicture}
	\caption{Using deep reinforcement learning, robust policies can be learned to achieve various tasks. Here, the reward function is shaped to encourage the throwing and catching of the pole after a rotation of $\SI{360}{\degree}$.}
	\vspace{-0.5cm}
	\label{fig:throw}
\end{figure}

In this paper, a strategy has been presented to transfer tactile control policies for the swing-up manipulation of different poles from simulation to a physical robotic system. As the simulator has been shown to closely match the dynamics of the real system, the policy learned in simulation generalizes to the real-world robotic system with no adaptation needed.
\hl{%
Note that the system presented here does neither exploit a fixed pivot point nor directly control the rotational degree-of-freedom of the pole, but it can achieve the desired motion only through the presence of friction, whose modeling was crucial in enabling a realistic simulation.%
}

This constitutes an important step towards a general framework to learn a wide variety of tactile manipulation tasks safely in simulation. Yet, current results have only been demonstrated for a single task on a single robotic system. Future work will focus on several aspects to further extend the generalizability of this work.
In a first step, the proposed framework could be utilized to learn other pole manipulation skills on the given system, e.g. the throwing and catching of a pole. While such a policy was already successfully learned in simulation (see \cref{fig:throw}), the transfer to the physical system requires further work due to non-idealities of the hardware. For instance, when the pole is thrown in the air, it may leave the plane to which the motion is assumed to be constrained.
In fact, instead of relying on the planar nature of the manipulation task, as was done in this work, the suggested simulator could be extended to handle non-planar tasks. As a result, manipulation skills for grippers that can be controlled in six degrees of freedom could also be learned.
Moreover, in this paper, hand-engineered features are extracted from the tactile observations, i.e. the orientation and total normal force acting on the pole. These features may not be relevant for other tasks, where learning such features end-to-end with the policy, e.g. using autoencoders, may further generalize the proposed framework.
%{\color{red} Mention something about auto-encoder approach?}
%exploits the planar nature of the swing-up manipulation task which may not be present for other tasks. As such , 

\addtolength{\textheight}{-0cm}   % This command serves to balance the column lengths
                                  % on the last page of the document manually. It shortens
                                  % the textheight of the last page by a suitable amount.
                                  % This command does not take effect until the next page
                                  % so it should come on the page before the last. Make
                                  % sure that you do not shorten the textheight too much.

%%%%%%%%%%%%%%%%%%%%%%%%%%%%%%%%%%%%%%%%%%%%%%%%%%%%%%%%%%%%%%%%%%%%%%%%%%%%%%%%

%%%%%%%%%%%%%%%%%%%%%%%%%%%%%%%%%%%%%%%%%%%%%%%%%%%%%%%%%%%%%%%%%%%%%%%%%%%%%%%%

%%%%%%%%%%%%%%%%%%%%%%%%%%%%%%%%%%%%%%%%%%%%%%%%%%%%%%%%%%%%%%%%%%%%%%%%%%%%%%%%
%\section*{APPENDIX}

%Appendixes should appear before the acknowledgment.

\section*{ACKNOWLEDGMENT}

The authors would like to thank M. Egli and M. Mueller for their contribution to the development of the system.

%%%%%%%%%%%%%%%%%%%%%%%%%%%%%%%%%%%%%%%%%%%%%%%%%%%%%%%%%%%%%%%%%%%%%%%%%%%%%%%%

\bibliographystyle{IEEEtran}
\bibliography{references}

\end{document}